\documentclass[]{article}
\usepackage[a4paper, total={6in, 8in}]{geometry}

\usepackage[square,numbers]{natbib}
\usepackage{wrapfig}
\usepackage[hidelinks]{hyperref}
\usepackage{authblk}

\usepackage{preamble}

\author[1]{Joshua Caiata}
\author[2]{Ben Armstrong\footnote{This work was primarily completed while affiliated with the University of Waterloo.}}
\author[1]{Kate Larson}

\affil[1]{University of Waterloo}
\affil[2]{Tulane University}
\affil[ ]{\texttt{jcaiata@uwaterloo.ca, research@benarmstrong.ca, kate.larson@uwaterloo.ca}}

\title{What Voting Rules Actually Do: A Data-Driven Analysis of Multi-Winner Voting}
\date{}

\begin{document}

\maketitle

\begin{abstract}

Committee-selection problems arise in many contexts and applications, and there has been increasing interest within the social choice research community on identifying which properties are satisfied by different multi-winner voting rules (e.g.~\cite{lackner2023multi}). In this work, we propose a data-driven framework to evaluate how frequently voting rules violate axioms across diverse preference distributions in practice, shifting away from the binary perspective of axiom satisfaction given by worst-case analysis. Using this framework, we analyze the relationship between multi-winner voting rules and their axiomatic performance under several preference distributions. We then show that neural networks, acting as voting rules, can outperform traditional rules in minimizing axiom violations. Our results suggest that data-driven approaches to social choice can inform the design of new voting systems and support the continuation of data-driven research in social choice.
 
\end{abstract}

\section{Introduction}

Committee selection is a central problem in social choice theory, wherein voters elect a committee (subset of alternatives) based on their preferences~\cite{lackner2023multi,faliszewski2017multiwinner}. 
There are numerous properties or axioms we might wish a multi-winner voting rule to satisfy, however many combinations of properties are known to be impossible to satisfy simultaneously.
Traditional research often focuses on ascertaining which axiomatic combinations are possible or are satisfied by a voting rule. Such research is motivated by a desire to identify rules which universally satisfy axioms deemed desirable for a particular setting. However, in many instances, a voting rule may not satisfy an axiom but rarely violates it in practice.

In this paper, we propose a data-driven framework to evaluate and explore voting rules axiomatically. To do so, we establish a measure of axiom violation more fine-grained than binary satisfaction and move away from worst-case analysis and towards an average case evaluation model. Using this framework, we explore the relationship between underlying voter preference distributions and axiomatic properties selected by common multi-winner voting rules using our average-case approach. We then use machine learning to learn bespoke multi-winner voting rules that aim to minimize axiom violations as much as possible when electing a committee. These multi-layer perceptrons are designed to select a set of alternatives based on preferences that violate certain axioms less frequently than many well-known rules.

Specifically, we make the following contributions:

\begin{enumerate}
    \item We initiate a data-driven measure of axiomatic violations for multi-winner voting rules.

    \item We explore the relationship between the underlying preference distributions and the axiomatic properties of committees selected by common multi-winner voting rules, illustrating the sensitivity of many of these rules to the underlying voter population.

    \item We empirically investigate how different multi-winner voting rules differ from each other both in terms of the committees they select and the frequency with which they violate (sets of) axioms.

    \item We demonstrate that it is possible to use machine learning to discover novel multi-winner voting rules that outperform existing rules under our evaluation framework.
    
\end{enumerate}
In summary, this paper illustrates the power of using data-driven approaches and machine learning to deepen our understanding of social choice and inform the design of new voting systems.

\subsection{Related Work}

Much prior work has both developed novel axioms to describe desirable properties, or shown that particular axioms are satisfied by certain voting rules.
Of relevance to our work,~\citeauthor{elkind2017properties} develop axioms for several types of multi-winner voting rules and establishes satisfaction results~\cite{elkind2017properties}. They note, in particular, the difficulty of satisfying Dummett's condition (where if a large enough group of voters agree on a set of alternatives as their top choices, then those alternatives should be in the winning committee) ~\cite{dummett1984}. A similar approach is taken for rules and axioms based on approval preferences \cite{lackner2023multi,peters2020proportionality}.

Our approach builds on data-driven axiomatic analysis~\cite{deon2020testing,fairstein2024learning}, drawing upon axioms from the social choice literature. Our results vary greatly based on underlying voter preferences; these are well-studied for the single-winner setting~\cite{elkind2018restricted} but less is known for the multi-winner setting. We use well-studied distributions shown to approximate human preferences or explore restricted cases~\cite{boehmer2021putting}.
Recent work has highlighted differences in the winners of multi-winner voting rules on generated and real-world preference data, emphasizing how some rules elect committees which are quite different from each other, particularly Minimax Approval Voting, Chamberlin-Courant, and sequential Chamberlin-Courant \cite{faliszewski2023experimental}.

Other recent work has explored the possibility of using machine learning with social choice. Prior work has primarily focused on approximating single-winner or probabilistic rules~\cite{matone2024deepvoting,burka2022voting,kujawska2020predicting}. Existing work on learning new voting rules, primarily in single-winner settings, has studied learning rules under specific axioms~\cite{armstrong2019machine}, reducing susceptibility to manipulation~\cite{firebanks2020machine}, or maximizing utility without axiomatic focus~\cite{anil2021learning}. One paper has explored the possibility of learning multi-winner rules for participatory budgeting with a focus predominantly on measures of social welfare \cite{fairstein2024learning}. Our work, instead, is focused on learning multi-winner rules aimed at minimizing axiom violations.

\section{Preliminaries: Social Choice Building Blocks}
\label{sec:prelim}

Traditional research in social choice is typically based upon three cornerstones: voting rules, voter preferences, and some measure of outcome quality.
Voting rules aggregate voter preferences following some particular procedure aimed at maximizing some measure of quality. Rules may provide a single winner, a ranking over alternatives, or a set of multiple winners.
Voter preferences are the subjects of aggregation; they are often modelled as originating from some fixed distribution or based upon data from real-world voting applications. Preferences might be expressed as rankings over alternatives, approvals, or even as raw scores.
Measures of voting rule quality primarily fall into one of two paradigms: \textit{social welfare}, or \textit{axiomatic}.

\begin{wrapfigure}[20]{r}{0.5\textwidth}
    \begin{center}
    \includegraphics[width=0.4\textwidth]{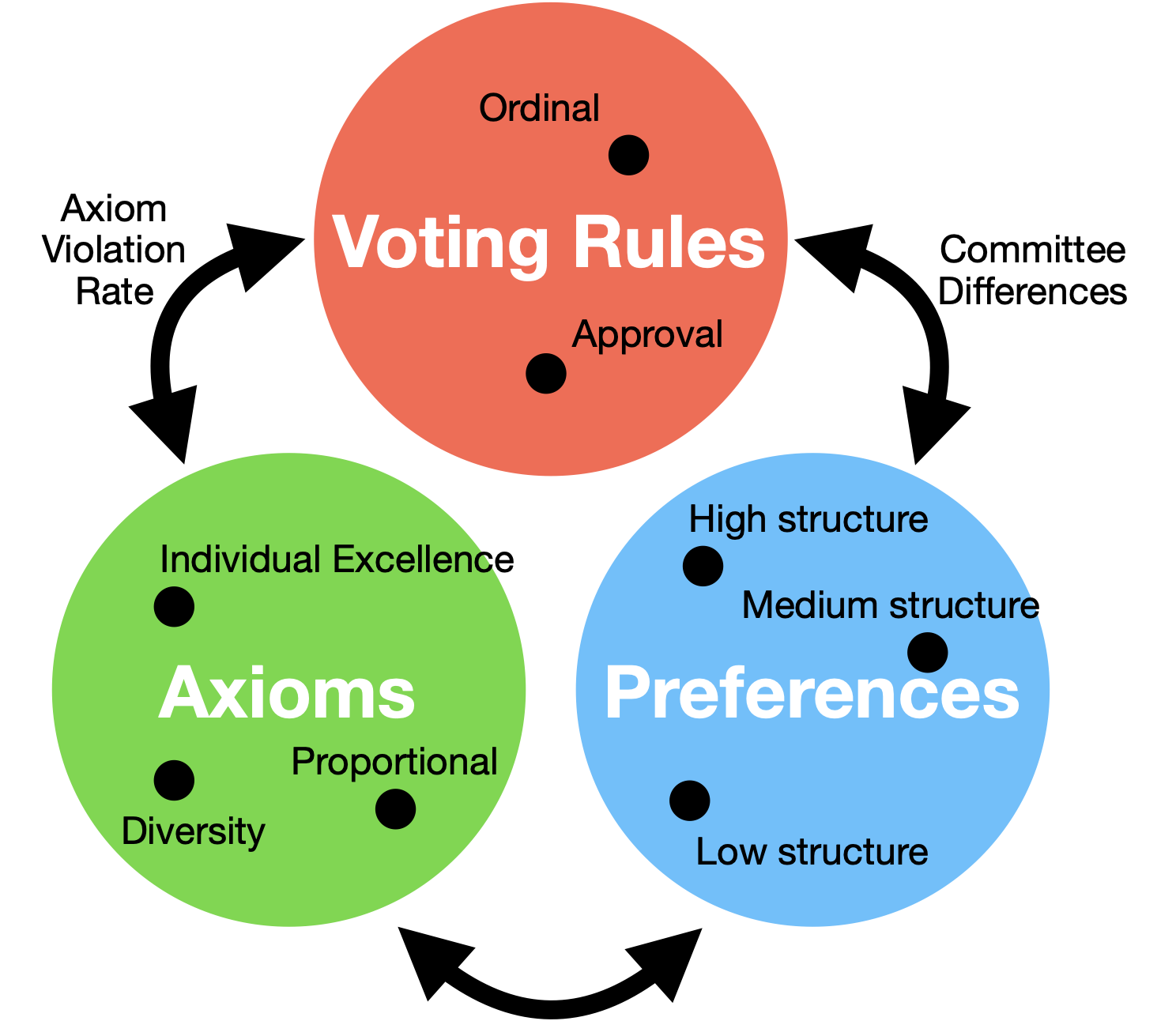}
    \end{center}
    \caption{An illustrated overview of the principle concepts in this paper. We apply an analytical framework to identify relationships between each of three fundamental cornerstones of social choice.}
\end{wrapfigure}

In this work we consider voting rules which elect multiple winners, using either ordinal or approval-based preferences sampled from a wide variety of distributions, and focus exclusively on a novel measure of the axiomatic properties of voting rules.
We now provide the basic notation used through the paper and describe each of these three cornerstones in more detail.

\subsubsection*{Model and Notation}

Let $V$ be a set of $n$ voters and $M$ be a set of $m$ alternatives.  Each voter $v_i\in V$ has a preference ranking over $M$ where for $a_i,a_j\in M$, $a_i\succ_v a_j$ means that voter $v\in V$ prefers alternative $a_i$ to $a_j$. A \emph{preference profile}, $P_\succ = (\succ_{v_1}, \ldots, \succ_{v_n})$ is a vector specifying the preferences of each voter.  In addition to voters' preferences, we are also interested in the alternatives that a voter approves. Let $App(v)\subset M$ be the \emph{approval set} of voter $v\in V$, containing the $k$ most preferred alternatives of $v$. We let $P_{App}=(App(v_1),\ldots, App(v_n))$, and when it is clear from the context we will abuse notation and let $P$ refer to either $P_\succ$ or $P_{App}$. 
An election, $E=(V, M)$, is defined by its voters and alternatives, along with, implicitly, either preference profiles or approval sets of the voters. 

We are interested in \emph{multi-winner} voting rules. Given an election $E$ and its associated $P$ and $k$,  $1 \leq k < m$, $\mathcal{F}_k(E)\subseteq \{C|C\subset M$, $|C| = k\}$ is a multi-winner voting rule that returns a family of $k$-sized subsets of $M$, called the winning \emph{committees}. 
If $\mathcal{F}_k$ directly uses $P_\succ$ we call the voting rule \emph{ordinal}. If $\mathcal{F}_k$ uses $P_{App}$ we say the voting rule is \emph{approval-based.}
We are interested in \emph{resolute} multi-winner voting rules and so assume that each rule uses some tie-breaking mechanism so that only one committee is returned. Unless otherwise stated we use lexicographic tie-breaking, sorting on the first non-shared alternative between committees.

\subsection{Multi-Winner Voting Rules}
\label{sec:existing_rules}

We consider multi-winner voting rules from the existing literature which are broadly classified into two categories: ordinal-based and approval-based. 
While much research focuses exclusively on just one of either ordinal- or approval-based rules we intentionally include rules from both categories in our study. This allows us to highlight behavioural differences among rules which are seldom compared directly.

The ordinal rules which we explore consist of:

\begin{description}
    \item[{\bf $k$-Borda}:] 
    Each voter assigns $m-1$ points to their top ranked alternative, $m-2$ points to their second ranked alternative etc. \rl{Borda} returns the $k$ alternatives with highest scores.

    \item[{\bf Single Non-Transferable Vote}:] 
    Each voter assigns one point to their most preferred alternative. \rl{SNTV} returns the $k$ alternatives with the highest points.\footnote{SNTV is an extension of plurality to the multi-winner context.}

    \item[{\bf Single Transferable Vote}:] \rl{STV} recursively adds to the set of winners all alternatives ranked first by voters with weight summing to more than a quota of $\frac{n}{k+1}$. Each voter begins with a weight of 1. When a winner, $a_w$ is found, each voter ranking $a_w$ first has their weight reduced by an amount proportional to the excess weight voting for $a_w$ beyond the quota. If there is not an alternative ranked first by a quota of weighted voters, the alternative ranked first by lowest sum of voter weight is removed from all voter rankings \cite{tideman1995single}.

\end{description}

We also consider the following approval-based rules:

\begin{description}

    \item[{\bf Bloc:}] 
     Define a score function $sc_\text{Bloc}(a)=\sum_{v\in V} \mathbf{1}_{\{a\in App(v)\}}$ giving one point to each alternative for each approval it receives. \rl{Bloc} returns the committee $C^*$ containing the $k$ alternatives receiving the most approvals.
     
     $$C^* = \arg\max_{\substack{C\subseteq M, \\|C|=k}} \sum_{a\in C} sc_\text{Bloc}(a)$$
    
     \item[{\bf Proportional Approval Voting: }]
     Given committee $C\subseteq M, |C|=k$, define 
     $$ sc_\text{PAV}(C)=\sum_{v\in V}\sum _{j=1}^{|C\cap App(v)|} \frac{1}{j}.$$
     
     \rl{PAV} returns $C^*=\arg\max sc_\text{PAV}(C)$. Each voter contributes some score to each committee based on the number of alternatives in the committee of which they approve; a voter approving of $r$ alternatives in $C$ adds the $r$-th harmonic number to the score of the committee.
    
     \item[{\bf Chamberlin-Courant (CC):}] 
     We consider three variations of the Chamberlin-Courant rule.  For committee $C\subset M, |C|=k$, define
     $$sc_{CC}(C)=\sum_{a\in C} \sum_{v\in V} \mathbf{1}_{\{ |C\cup App(v)|>0\}}.$$
     $sc_{CC}(C)$ is the number of voters who approve at least one alternative in $C$.
     
     \noindent
     {\bf Approval CC} is defined as \rl{CC} $ = C^*$ where $C^*=\arg\max sc_{CC}(C)$.   

     \noindent
     {\bf Lexicographic CC}, \rl{lex-CC}, maximizes $sc_{CC}$, breaking ties by selecting the committee maximizing the number of voters approving 2 alternatives (then, if ties remain, 3 alternatives, etc.).

     \noindent
     {\bf Sequential CC}, {seq-CC}, constructs a winning committee by iteratively adding $a\in M$ that increases the $sc_{CC}$ the most at each step.

     \item[{\bf Monroe: }] 
     Considers all ways of assigning each voter one alternative in committee $W$, such that every $a \in W$ is assigned to between $\floor{\frac{n}{k}}$ and $\ceil{\frac{n}{k}}$ voters. The score of an assignment is the sum of Borda scores of each voter's assigned alternative.
     \rl{Monroe} selects the committee with the highest score. 

     \noindent
     \textbf{Greedy Monroe}, \rl{Greedy M.}, constructs a winning committee iteratively following a similar scoring process\footnote{We  refer the reader to ~\citeauthor{lackner2023multi} for a formal definition \cite{lackner2023multi}.}. 
    
    \item[{\bf Minimax Approval:}]
     \rl{MAV} selects the committee that minimizes the maximum Hamming distance between any voter's approved alternatives and the committee. 

     \item[\textbf{Method of Equal Shares:}] \rl{MES} has two phases. First, each voter has a budget of $\frac{k}{n}$. Proceed for up to $k$ rounds: Adding a voter to the committee has a cost of $1$, which can be split between many voters. In each round, consider alternatives $A_r$ which are not in the committee and are approved of by voters that have a remaining budget summing to at least 1. If $A_r$ is empty, go to phase 2.
    Otherwise, select $a \in A_r$ such that each voter approving of $a$ must spend at most $\rho$ to add them to the committee. Add $a$ to the committee, adjust the remaining budget of each voter, and proceed to the next round.
    In the second phase, many possible rules can be used to fill any remaining spots on the committee. We use the sequential Phragmen rule. MES satisfies the JR and PJR axioms (but not the Core). See \citeauthor{lackner2023multi} for further details \cite{lackner2023multi}.

    \item[\textbf{E Pluribus Hugo:}] Also called ``Single Divisible Vote with Least-Popular Elimination''; \rl{EPH} operates in rounds. In each round, each voter divides a single point evenly between all remaining alternatives of which they approve. Alternatives are ranked in order of total summed points from all voters. The two alternatives with the lowest number of points are compared: the one receiving the fewest approvals overall is eliminated. Rounds of elimination continue until $k$ alternatives remain. \rl{EPH} satisfies the Strong Pareto Efficiency axiom \cite{quinn2019proportional}. 

    \item[\textbf{Random Serial Dictator:}] \rl{RSD} selects a single voter to serve as ``dictator''. The winning committee is exactly the set of alternatives approved of by that voter.
    
\end{description}

\citeauthor{faliszewski2017multiwinner} describe three categories for multi-winner voting rules: \textit{individual excellence} (electing alternatives which are individually well-liked), \textit{diversity} (electing alternatives which are different from each other), and \textit{proportionality} (electing a committee which proportionally represents the preferences of voters). The rules we use are generally aligned with one or two of these categories \cite{faliszewski2017multiwinner,faliszewski2023experimental}:

\begin{description}

    \item[ \textbf{Individual Excellence:}] \rl{Borda}, \rl{SNTV}, \rl{Bloc}, \rl{EPH}
    \item[ \textbf{Diversity:} ] \rl{SNTV}, \rl{CC}
    \item[ \textbf{Proportionality:}] \rl{STV}, \rl{PAV}, \rl{Monroe}, \rl{CC}, \rl{MES}, \rl{EPH}
\end{description}

We categorize the variants of each rule in the same way as the original rule. We highlight that these are subjective categorizations and are not mutually exclusive. For example, $\mathcal{F}^{CC}$ has been described as both \textit{diverse} \cite{faliszewski2023experimental} and \textit{proportional} \cite{elkind2017properties}.
For this reason we generally group the two categories together.
Additionally, $\mathcal{F}^{MAV}$ and $\mathcal{F}^{RSD}$ do not neatly fit into any category ($\mathcal{F}^{MAV}$ considers alternatives as sets, rather than individuals, but does not obviously aim to achieve diversity or proportionality while $\mathcal{F}^{RSD}$ considers only a single voter's opinion).

\subsection{Voting Rule Axioms}

Much of the literature on voting rules is axiomatic, describing desirable properties that voting rules may or may not exhibit. We focus exclusively on \textit{intraprofile} axioms -- axioms for which we can determine a violation using only the preference profile being given to $\mathcal{F}$ and the resulting committee \cite{schmidtlein2022voting}. Axioms can also be loosely categorized based on the stated priorities of their definitions. 

We use the following axioms which describe aspects of \textit{individual excellence}:

\begin{description}

\item[\textbf{Majority Winner}]  If $\ceil{\frac{n}{2}}$ or more voters rank alternative $x$ first in their ballot, then $x$ is in the winning committee~\cite{fishburn1977condorcet}.

\item[\textbf{Majority Loser}]  If $\ceil{\frac{n}{2}}$ or more voters rank alternative $x$ last in their ballot, then $x$ is not in the winning committee.

\item[\textbf{Condorcet Winner}]  A Condorcet committee $C$ is one in which for all $x\in C$ and for all $y\in M\setminus C$,  the majority of voters prefer $x$ to $y$. A voting rule satisfies the Condorcet Winner axiom if, whenever one exists, it returns a Condorcet committee~\cite{gehrlein1985condorcet}.

\item[\textbf{Condorcet Loser}] Let $L\subseteq M$ such that for all $x\in L$, and $y\not\in L$, $y$ is preferred to $x$ by a majority of voters. A voting rule satisfies the Condorcet Loser axiom if it never returns $L$~\cite{gehrlein1985condorcet}.

\item[\textbf{Strong Pareto Efficiency}]  Committee $C$ dominates $C'$ if every voter approves at least as many alternatives in $C$ as in $C'$, and at least one voter approves strictly more alternatives in $C$ than in $C'$. A multi-winner voting rule exhibits Strong Pareto Efficiency if it never returns a dominated committee~\cite{lackner2023multi}.

\item[\textbf{Fixed Majority}] If there exists a set of  alternatives $C,\ |C|=k$ and a set of voters $X \subseteq V$ with $|X| > \frac{n}{2}$ that all rank each alternative in $C$ above each alternative not in $C$ then the winning committee is $C$~\cite{debord1993prudent,elkind2017properties}.

\item[\textbf{Strong Unanimity}]  If every voter ranks the same $k$ alternatives on top, then those alternatives form the winning committee~\cite{elkind2017properties}.

\end{description}

And we use the following axioms that prioritize \textit{diversity} or \textit{proportionality}.

\begin{description}

\item[\textbf{Dummett's Condition}]  If there is a group of $\frac{\ell \cdot \textit{n}}{\textit{k}}$ voters that all rank the same $\ell$ alternatives on top, these $\ell$ alternatives are in the winning committee~\cite{dummett1984}. Also referred to as ``Proportionality for Solid Coalitions'' in some literature \cite{brill2023robust}.

\item[\textbf{Local Stability}] For $q = \ceil{\frac{n}{k}}$, committee $C$ violates local stability if there exists a subset of voters $V^*$ with $|V^*| \geq q$ and an alternative $x \notin C$ such that every voter in $V^*$ prefers $x$ to all members of $C$~\cite{aziz2017condorcet}. We highlight in particular that a locally stable committee is \textit{not} guaranteed to exist. That is, it is possible for there to exist a preference profile for which any winning committee would violate local stability.

\item[\textbf{Solid Coalitions}]  If at least $\frac{n}{k}$ voters rank some alternative $x$ first, then $x$ should be in the winning committee~\cite{elkind2017properties}. 

\item[\textbf{Core}]  A committee $C$ is ``in the core'' if for each non-empty subset of $\mathcal{V} \subseteq V$ of voters, and each non-empty subset of alternatives $T \subseteq A$ with 

\[
\frac{|T|}{k} \leq \frac{|V|}{n}
\]

there is a $v_i \in V$ such that $|App(v_i) \cap T| \leq |App(v_i) \cap W|$. That is, $v_i$ approves of at least as many alternatives in $W$ as they approve of in $T$. A voting rule satisfies the Core if it always returns a committee that is in the core \cite{lackner2023multi}.

\item[\textbf{(Extended) Justified Representation}]
For $\ell \geq 1$, a group of voters is $\ell$-cohesive if (1) $|V| \geq \ell \cdot \frac{n}{k}$, and (2) $|\bigcap_{i \in V} App(v_i)| \geq \ell$.

A voting rule satisfies Justified Representation \textbf{(JR)} if for a winning committee $C$, it is always the case that every $1$-cohesive group of voters $\mathcal{V}$ contains a $v_i$ who approves of at least one member of $C$; i.e. $|W \cap App(v_i)| \geq 1$.

A voting rule satisfies Extended Justified Representation \textbf{(EJR)} if for a winning committee $C$, it is always the case that  \textit{every} $\ell$-cohesive group of voters $\mathcal{V}$ contains a $v_i$ that approves of at least $\ell$ winners, for $1 \leq \ell \leq k$ \cite{lackner2023multi}.

\end{description}

\autoref{tab:summary_results} illustrates which existing voting rules are known to satisfy each axiom. Green entries indicate known axiomatic satisfaction for a particular axiom and voting rule.

The axioms we use are only one possible set and were chosen to represent a wide variety of desirable properties.
While our framework is general and applies to \textbf{any} set of intra-profile axioms, we include a study of the relationships between these axioms in \Cref{app:axiom_implications}. While relationships between some of our axioms are already described in the literature, we also identify several novel relationships where the satisfaction of one axiom implies the satisfaction of another.

\subsection{Preference Distributions}
\label{sec:pref_distributions}

We use a wide range of standard preference distributions ($\mathcal{D}$) through our experiments. These are used both for training novel multi-winner rules, as well as testing novel and existing rules.
Note that we generate ordinal ballots and, when needed by a particular rule, convert them to an approval-based format where each voter approves of their $k$ top-ranked alternatives. There are many other valid methods of generating approval ballots which we do not explore.

Through our experiments we consider 8 unique families of preference distribution, as well as two additional sets of preferences. We loosely categorize each of these distributions by the amount of structure within the preferences they generates, from Identity preferences where all voters are identical to Impartial Culture where voters are assigned preferences uniformly at random.

\paragraph{Unstructured Distributions}

\begin{itemize}
    \item[] \textbf{Impartial Culture:} Each voter draws its preference ranking from a uniform distribution over all possible rankings~\cite{guilbaud1952theories}.
    \item[] \textbf{Impartial Anonymous Culture:} Given $n$ voters, each $n$-element multiset of preference rankings is drawn from a uniform distribution~\cite{gehrlein1976condorcet}.
\end{itemize}

\paragraph{Moderately Structured Distributions}

\begin{itemize}
    \item[] \textbf{Mallows:} Given reference ranking $\sigma$ and dispersion parameter $\theta$, the probability of drawing ranking $r$ is $P(r|\sigma, \theta)=\frac{1}{Z}\theta^{d(r,\sigma)}$ where $Z$ is a normalization factor and $d(r,\sigma)$ is the Kendall-Tau distance between rankings $r$ and $\sigma$~\cite{mallows1957non,boehmer2021putting}.
    \item[] \textbf{Urn}: Parameterized by $\alpha$. All $m!$ preference orders exist in an ``urn.'' Voters select orders consecutively. After a preference order is selected, it is returned to the urn along with $\alpha!$ copies of the ranking \cite{eggenberger1923statistik}.
    \item[] \textbf{Euclidean:} Parameterized by a dimension and a topology. Voters and alternatives are placed randomly within the chosen space and preferences correspond to the distance from each voter to each alternative \cite{enelow1984spatial}.
    \item[] 
\end{itemize}

\paragraph{Highly Structured Distributions}

\begin{itemize}
    \item[] \textbf{Identity:} All voters in $V$ have identical preferences.
    \item[] \textbf{Single-Peaked:} A global ordering of alternatives exists; each voter has a favourite alternative and prefers alternatives closer to their favourite over those further ~\cite{black48rationale}.
    \item[] \textbf{Stratification:} Parameterized by a weight $w \in (0, 1)$. Alternatives are split into two classes with the first class's size proportional to $w$.
    All voters rank all alternatives in the first class above those in the second class. 
    Within a class voters rank alternatives uniformly at random~\cite{boehmer2021putting}.
\end{itemize}

\paragraph{Additional Preference Distributions}

\begin{itemize}
    \item[] \textbf{Mixed:} In addition to preference distributions taken on their own, we consider a distribution consisting of profiles sampled from every other distribution with equal probability.

    \item[] \textbf{PrefLib:} We include all applicable strict, complete preference profiles from PrefLib, an online repository of human preference data \cite{mattei2013preflib}. The amount of PrefLib data applicable to our setting is not sufficient for learning from; we use this data exclusively during evaluation.
\end{itemize}

\section{Data-Driven Voting Rule Analysis}
\label{sec:axiom}

Traditionally, axiomatic analysis of voting rules asks whether a rule satisfies some axiom universally, which is a binary question. In the real world and synthetic preference profiles, however, the gap between worst-case analysis and average-case behaviour can be quite large. Rules that are known to explicitly not satisfy an axiom may very rarely violate such an axiom in practice. The key insight, and motivation, for such a shift in analyzing a voting rule is that a fine-grained and empirical lens reveals real and meaningful information on how rules behave across distributions, which are not considered when only looking at axiomatic satisfaction as a binary question.

We introduce two measures that we use to understand the behaviour of voting rules. The data-driven analysis we employ is very well-suited to exploring deep non-binary measures of axiom satisfaction. We capture this by measuring the \textit{rate} at which the outcome of voting rules violates an axiom. We also study the amount of overlap between committees elected by voting rules. This provides us an indication of how similar two rules in practice, regardless of how they function internally.

While both of our data-driven metrics are affected by each of the three cornerstones of our social choice framework, they are more naturally associated with certain components. The axiom violation rate captures the interplay between axiomatic performance and voting rule behaviour\footnote{We argue that preferences are less intrinsically related to axiom violations than our other cornerstones: in the cases where a voting rule satisfies a particular axiom then the underlying preference distribution has no bearing on the violation rate.} while the rule difference measure reveals connections between preference distributions and voting rules\footnote{This is because axioms have no direct bearing on the output of any given voting rule -- although they may indirectly influence the design of voting rules.}.

\subsection{Axiom Violation Rate}

We begin by formalizing the axiom violation rate (AVR) as our core metric for empirical axiomatic performance. As we focus exclusively on \textit{intraprofile axioms}, we can determine whether an axiom is violated by a preference profile using only the voting rule $\mathcal{F}$ and the profile itself.

We now define a measure of the frequency that an axiom is violated by a voting rule on given preference profiles.
If axiom $A$ is violated by a specific preference profile $P$ and committee $c$, we say $A(P, c) = 1$. If $A$ is not violated, $A(P, c) = 0$. Then, we define the axiom violation rate (AVR) of $\mathcal{F}$ on a set of profiles $\mathbb{P}$ over axioms $\mathbb{A}$ as:

\[
\text{AVR}(\mathcal{F}, \mathbb{P}, \mathbb{A}) = \frac{1}{|\mathbb{A}||\mathbb{P}|} \sum_{A \in \mathbb{A}} \sum_{P \in \mathbb{P}}  A(P, \mathcal{F}(P))
\]

\subsection{Rule Differences}

We also measure overlap between elected committees. This tells us (1) whether rules with similar AVR elect similar underlying committees, and (2) the degree of similarity between committees elected by rules with differing AVRs.
We say $\cap^+_P(\mathcal{F}^1, \mathcal{F}^2) = \mathcal{F}^1(P) \cap \mathcal{F}^2(P)$
and
$\cap^-_P(\mathcal{F}^1, \mathcal{F}^2) = (M \setminus \mathcal{F}^1(P)) \cap (M \setminus \mathcal{F}^2(P))$. We also define a normalization factor $\delta = \frac{m}{m-|m-2k|}$ which ensures the difference between two rules on a given set of profiles $\mathbb{P}$ has a range from 0 to 1.

\[
\texttt{d}(\mathcal{F}^1, \mathcal{F}^2, \mathbb{P}) = \delta -\frac{\delta}{m|\mathbb{P}|} \sum_{P \in \mathbb{P}} |\cap^+_P(\mathcal{F}^1, \mathcal{F}^2)| + |\cap^-_P(\mathcal{F}^1, \mathcal{F}^2)|
\]

\section{Learning Rules to Satisfy Axioms}

We now describe \rl{NN}, a novel voting rule with a functionality directly based upon goals rooted in our cornerstones of axiom satisfaction and voter preferences. Functionally, \rl{NN} is multi-layer perceptron trained to predict a winning committee based on a given preference profile. This section outlines the procedure we use to train the model. 
Specific details on the parameters used to train \rl{NN} in our experiments are included in \autoref{sec:experiments}.

\subsection{Generating Axiom Violation Data}

We generate separate training sets (used by \rl{NN}) and testing sets (used to evaluate all rules). A single example in a dataset is generated as follows:

\begin{enumerate}
    \item Given some preference distribution, $\mathcal{D}$ (see \autoref{sec:pref_distributions}) sample a preference profile $P$ from $\mathcal{D}$
    \item Randomly rename each alternative in $P$.\footnote{Renaming alternatives ensures that our data satisfies neutrality by treating alternatives identically. Otherwise, some preference distributions are biased towards specific outcomes. For example, our \textit{identity} distribution only generates the ranking $a_1 \succ a_2 \succ ...$ which will bias a network to always elect $a_1$ through $a_k$.}
    \item Find, by exhaustive search, the committee $c = \argmin_c \sum_{A \in \mathbb{A}} A(P, c)$ minimizing the number of axiom violations in $\mathbb{A}$ with ties broken lexicographically.
\end{enumerate}

\subsection{Input Data Structure}

We transform each preference profile into three separate matrices, preserving aspects of voter preferences while reducing input size to depend only on $m$, the number of alternatives. This ensures models are applicable for any number of voters $n$, while being specific to $m$ and committee size $k$.

\textbf{Majority Matrix}
The preference profile is transformed into an $m \times m$ matrix $\mathcal{R}^\text{majority}$ with $\mathcal{R}_{ij}^\text{majority} = 1$ if a weak majority of voters prefer $a_i$ to $a_j$ and 0 otherwise.

\textbf{Weighted Preference Matrix}
The preference profile is transformed into an $m \times m$ matrix $\mathcal{R}^\text{weighted}$ with $\mathcal{R}_{ij}^\text{weighted} = c$ to indicate that $c$ voters prefer $a_i$ over $a_j$.

\textbf{Ranking Matrix}
The preference profile is transformed into an $m \times m$ matrix $\mathcal{R}^\text{ranked}$ where $\mathcal{R}_{ij}^\text{ranked} = c$ indicates that $c$ voters rank $a_i$ in position $j$.

The majority and weighted preference matrices correspond, respectively, to tournaments and weighted tournaments in the literature~\cite{brandt2016handbook}. We flatten these matrices to 1-dimensional arrays. $\mathcal{R}^\text{weighted}$ and $\mathcal{R}^\text{ranked}$ are stored in a normalized form so that the model output is agnostic of the exact number of voters present. We concatenate each of these arrays to form the input to our models. To simplify notation, we continue to write $\mathcal{F}(P)$ when discussing the output of a trained model; however, in place of $P$, the function receives only the above transformations of $P$.

\subsection{Model Loss}

During training, committees of input examples are encoded as $k$-hot vectors: e.g. [0, 1, 1, 0, 1] indicates that alternatives 2, 3, and 5 are in the winning committee while alternatives 1 and 4 are not in the winning committee.
Our model uses the L1 loss function\footnote{We tested all applicable loss functions in Pytorch and found that L1 Loss provided the best results~\cite{paszke2019pytorch}.}, defined as:

\[L_1(\mathbf{y}, \hat{\mathbf{y}}) = \frac{1}{n} \sum_{i=1}^{n} \left| y_i - \hat{y}_i \right|\]
where $y$ is the committee from the training example and $\hat{y}$ is the model's predicted output.
As the model does not output integer values, the output must be processed to form a committee: the highest $k$ values in the length $m$ list output by the model are set to $1$ while the other values are set to $0$.

\section{Experimental Results and Analysis}
\label{sec:experiments}

We now describe the experiments we have run, provide their results and discuss their implications. From both existing and learned rules we are able to extract novel conclusions about each of our cornerstones and the learnability of novel rules.

\begin{figure*}[t]
    \centering
    \includegraphics[width=\linewidth]{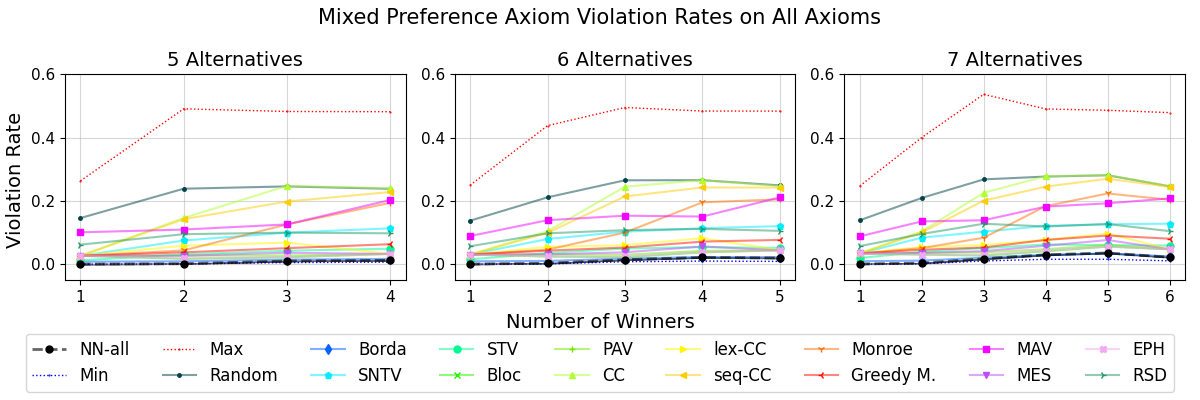}
    \caption{Axiom violation rate averaged over all axioms for each number of voters and winners on Mixed preferences.
    }
    \label{fig:mixed-distribution}
\end{figure*}

\begin{figure*}[t]
    \includegraphics[width=\textwidth]{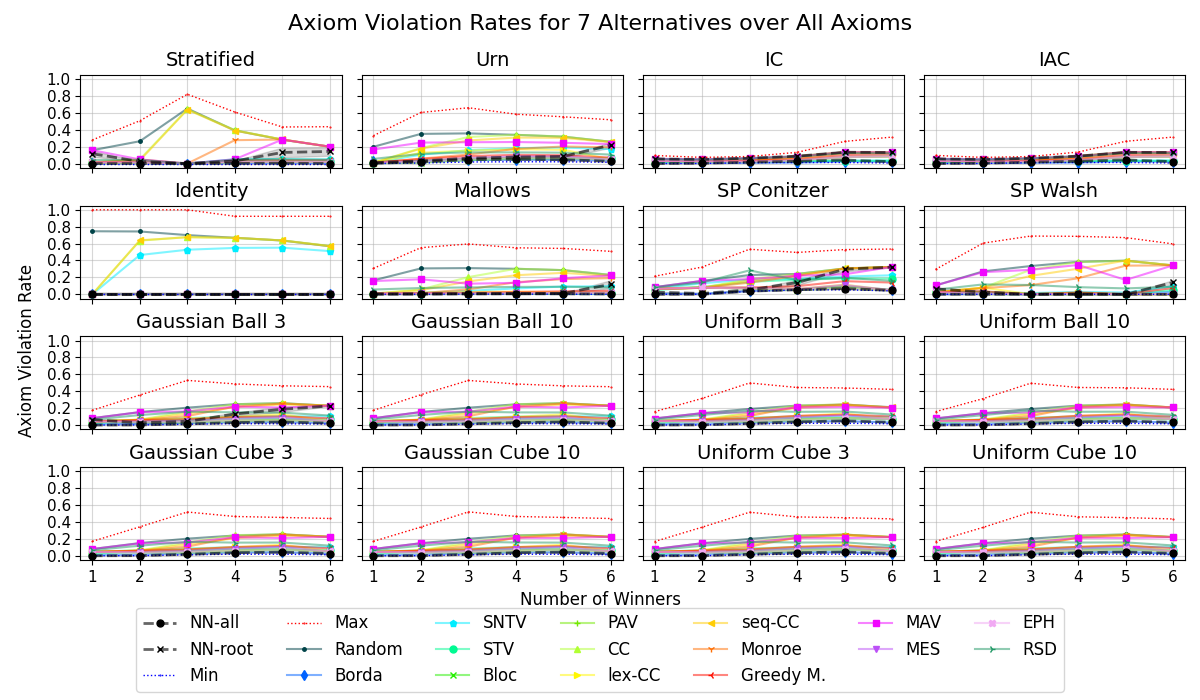}
    \caption{Axiom violation rates for each rule under each individual preference distribution for $m=7$. In all cases, $\mathcal{F}^\text{NN-all}$ has AVR lower than, or similar to, other rules.
    }
    \label{fig:violations_by_distribution-m=7}
\end{figure*}

\subsection{Experimental Training Parameters}
\label{sec:learning_params}

In our experiments we learn two different rule configurations. These configurations differ primarily in which axioms are evaluated in the loss function.

\paragraph{Learning Configuration 1: All Axioms} Our initial learned rule \rl{NN-all} is trained on data generated from evaluating violations for \textit{all} axioms we have described. We train \rl{NN-all} on all test distributions described below.

\paragraph{Learning Configuration 2: Root Axioms} We also train \rl{NN-root} on a second set of axioms: those axioms which, if satisfied, imply the satisfaction of all other axioms we use. This includes: Majority Loser, Condorcet Winner, Dummett's Condition, Local Stability, Strong Pareto, and the Core. See \autoref{app:axiom_implications} for details of these axiom relationships. We train \rl{NN-root} on all test distributions described below, except as noted.

\subsubsection{Preference Distributions}

Across both configurations most parameter values are held constant. We train (as applicable) and test all rules (learned and existing) on all combinations of $m = \{5, 6, 7\}$ alternatives and $k = \{1, 2, ..., m-1\}$ winners. In all cases we use profiles with $n=50$ voters.
All rules are evaluated on 25,000 profiles sampled from each of the following test distributions with learned rules being tested on the distribution upon which they were trained.

\begin{itemize}
    \item IC, IAC, Identity take no parameter
    \item Mallow's with $\theta$ sampled uniformly at random as described by \citeauthor{boehmer2023properties}\cite{boehmer2023properties}.
    \item Urn with $\alpha$ sampled from a Gamma distribution \cite{boehmer2021putting}.
    \item Single-peaked distributions as described by Conitzer~\cite{conitzer2007eliciting} and Walsh~\cite{walsh2015generating}. 
    \item Stratification with $w=0.5$.
    \item 8 Euclidean distributions with each combination of: 3 or 10 dimensions, a Ball or Cube topology, and Uniform or Gaussian placement of voters. Note: Due to the similarity of results across Euclidean distributions from \rl{NN-all}, we trained \rl{NN-root}only on one Euclidean distribution, the 3-dimensional Gaussian Ball.
    \item 1 distribution containing an even mixture of the 16 other distributions. We test mixed distributions on \rl{NN-all} and pre-existing rules, but not \rl{NN-root}.
\end{itemize}

\subsubsection{Learning Parameters}

For each configuration, we train 20 neural networks using PyTorch \cite{paszke2019pytorch}, with 5 hidden layers of 256 nodes. Networks are trained for up to 50 epochs using the Adam optimizer with a learning rate of 0.0001, stopping early if no improvement of 0.0005 occurs over 10 epochs.
We generate separate training and testing data sets of 25,000 examples each for all 255 unique combinations of $m$, $k$, and $\mathcal{D}$.
Each profile contains 50 voters\footnote{In preliminary experiments we have trained networks using data containing a number of voters sampled from a normal distribution truncated between 25 and 75 voters. We found no difference to performance and use profiles with 50 voters for simplicity.}. We do not filter profiles to ensure no overlap between training and test sets but find that there is minimal overlap between these sets in almost all cases. \footnote{In the identity distribution all profiles are identical. On profiles sampled from the Urn distribution for 5, 6, and 7 alternatives roughly 3.5\%, 2\%, and 0.5\% of profiles overlapped on a training distribution of 25,000 profiles. For Mallows preferences with 5 alternatives, up to 0.2\% of profiles overlap. In all other cases training distributions have zero overlap.}

\begin{table}[ht]
\centering
\fontsize{7pt}{9pt}
\selectfont
\setlength{\tabcolsep}{4.6pt}
\renewcommand{\arraystretch}{1.05}\begin{tabular}{lc|cccccc|ccccccc}
\toprule
Method & \rotatebox{90}{Mean} & \rotatebox{90}{Maj W} & \rotatebox{90}{\underline{Maj L}} & \rotatebox{90}{\underline{Cond W}} & \rotatebox{90}{Cond L} & \rotatebox{90}{\underline{Pareto}} & \rotatebox{90}{F Maj} & \rotatebox{90}{Unanimity} & \rotatebox{90}{\underline{Dummett's}} & \rotatebox{90}{JR} & \rotatebox{90}{EJR} & \rotatebox{90}{\underline{Core}} & \rotatebox{90}{S. Coalitions} & \rotatebox{90}{\underline{Stability}} \\
\midrule
NN-all & .017 & \textit{.000} & \textit{.000} & \textbf{.015} & \textit{.000} & .004 & \textit{.000} & \textit{.000} & .061 & .001 & .001 & .001 & .046 & .092 \\
NN-root & .038 & .001 & .030 & .200 & .010 & .045 & .024 & \textit{.000} & .056 & \textit{.000} & \textit{.000} & .002 & .044 & .082 \\
Min & \textbf{.009} & \textbf{0} & .001 & .036 & \textbf{0} & .001 & .001 & \textbf{0} & .012 & \textbf{0} & \textit{.000} & \textit{.000} & .005 & .059 \\
Max & .440 & .125 & .340 & .919 & .635 & .620 & .175 & .076 & .555 & .234 & .354 & .381 & .521 & .787 \\
\midrule
Borda & .021 & .001 & .004 & .125 & \textbf{0} & .004 & .011 & \cellcolor{green!25}\textbf{0} & .044 & \textit{.000} & \textit{.000} & \textit{.000} & .031 & .056 \\
EPH & .040 & \textit{.000} & .001 & .270 & .002 & \cellcolor{green!25}\textit{.000} & .001 & \textbf{0} & .082 & \textit{.000} & \textit{.000} & \textit{.000} & .063 & .096 \\
SNTV & .099 & \cellcolor{green!25}\textbf{0} & .098 & .619 & .007 & .227 & .106 & .049 & .062 & .001 & .054 & .058 & \cellcolor{green!25}\textbf{0} & .012 \\
Bloc & .039 & \textit{.000} & .001 & .254 & .002 & \cellcolor{green!25}\textbf{0} & \cellcolor{green!25}\textbf{0} & \cellcolor{green!25}\textbf{0} & .080 & \textit{.000} & \textit{.000} & \textit{.000} & .061 & .106 \\
\midrule
STV & .048 & \cellcolor{green!25}\textbf{0} & .037 & .442 & .002 & .118 & .029 & \textbf{0} & \textbf{0} & \textit{.000} & \textit{.000} & .001 & \cellcolor{green!25}\textbf{0} & \textbf{.001} \\
PAV & .043 & .001 & .001 & .308 & .002 & \cellcolor{green!25}\textbf{0} & .004 & \textbf{0} & .088 & \cellcolor{green!25}\textbf{0} & \cellcolor{green!25}\textbf{0} & \textbf{0} & .068 & .091 \\
MES & .049 & .001 & .002 & .351 & .002 & .001 & .008 & \textbf{0} & .096 & \cellcolor{green!25}\textbf{0} & \cellcolor{green!25}\textbf{0} & \textbf{0} & .075 & .095 \\
CC & .195 & .036 & .146 & .756 & .031 & .344 & .141 & .062 & .308 & \cellcolor{green!25}\textbf{0} & .084 & .091 & .232 & .301 \\
seq-CC & .183 & .032 & .139 & .740 & .025 & .297 & .140 & .061 & .292 & \cellcolor{green!25}\textbf{0} & .078 & .081 & .216 & .278 \\
lex-CC & .061 & .005 & .007 & .440 & .002 & \textbf{0} & .024 & \textbf{0} & .117 & \textbf{0} & \textit{.000} & \textit{.000} & .091 & .112 \\
Monroe & .130 & .007 & .078 & .649 & .026 & .234 & .060 & \cellcolor{green!25}\textbf{0} & .214 & \cellcolor{green!25}\textbf{0} & .002 & .006 & .180 & .231 \\
Greedy M. & .063 & .002 & .019 & .448 & .003 & .012 & .023 & \cellcolor{green!25}\textbf{0} & .112 & \cellcolor{green!25}\textbf{0} & \textbf{0} & \textbf{0} & .089 & .118 \\
\midrule
MAV & .157 & .022 & .110 & .750 & .044 & .279 & .084 & \textbf{0} & .219 & .015 & .022 & .022 & .179 & .300 \\
RSD & .105 & .008 & .056 & .594 & .016 & \textbf{0} & .036 & \textbf{0} & .148 & .030 & .032 & .033 & .120 & .299 \\
Random & .237 & .063 & .171 & .845 & .057 & .406 & .160 & .071 & .326 & .049 & .125 & .134 & .252 & .419 \\
\bottomrule
\end{tabular}
\caption{
Axiom violation rates for 7 alternatives averaged over all distributions and numbers of winners. Voting rules and axioms are separated to indicate to which category of rule/axiom they belong. Root axioms are underlined. Bold values indicate the best result of a column, italic values have been rounded to zero. Shaded green indicates that previous work has shown the rule satisfies this axiom \protect\cite{elkind2017properties,lackner2023multi}. Due to differences in tie-breaking with previous work, some edge cases do not match prior theoretical results.
}
\label{tab:summary_results}
\end{table}

\subsection{Understanding Our Cornerstones}

We first examine what we can observe by considering each pair of our three cornerstones: axioms, voting rules, and preferences. By examining multiple aspects of our area of focus we are able to form novel connections between each topic.

\subsubsection{Axioms and Voting Rules}

In \autoref{tab:summary_results} we show the violation rate of each voting rule on each individual axiom, as well as an average violation rate over all axioms. This allows us to directly explore the relationship between voting rules and axiomatic performance. After dividing rules and axioms into their informal categories of excellence-based or diverse/proportional some consistent patterns emerge.

Excellence based rules generally have quite low axiom violation rates. Being good at excellence-based axioms  leads to strong performance on diverse/proportional axioms. 
The axioms that are formally satisfied by some proportionality-based rules (i.e., JR, EJR) are almost never violated in practice by most excellence-based rules. Other proportionality-based axioms, such as Solid Coalitions, are violated more frequently by all diverse/proportional rules except \rl{STV}.

\textul{\textbf{Observation:}} Alternatives which are liked individually are also likely to be members of committees which provide strong proportional properties. Identifying the alternatives which are liked individually may be easier to do well at than identifying strong sets of alternatives.

\textul{\textbf{Observation:}} Certain axioms -- Condorcet Winner, Dummett's Condition, Local Stability -- are significantly harder to avoid violating than others. This pattern is consistent across all rules and is also reflected in the violation rate of the best committees on these axioms.

\textul{\textbf{Observation:}} Learning a rule to minimize axiom violation rate results in strong performance across \textit{all} axioms. Curiously, the learned rules violate the Condorcet Winner condition at a rate lower than they would were they to perfectly minimize violations across all axioms.

\textul{\textbf{Observation:}} Learning basic axioms can inform us about more complex axioms. \rl{NN-root}, the rule trained to minimize violations on only the root axioms, has an average violation rate lower than most rules but much higher than \rl{NN-all}. Notably, \rl{NN-root} has a higher violation rate \textit{on most root axioms} than \rl{NN-all}.

\subsubsection{Axioms and Preference Distributions}

\begin{figure}[ht]
    \centering
    \includegraphics[width=\linewidth]{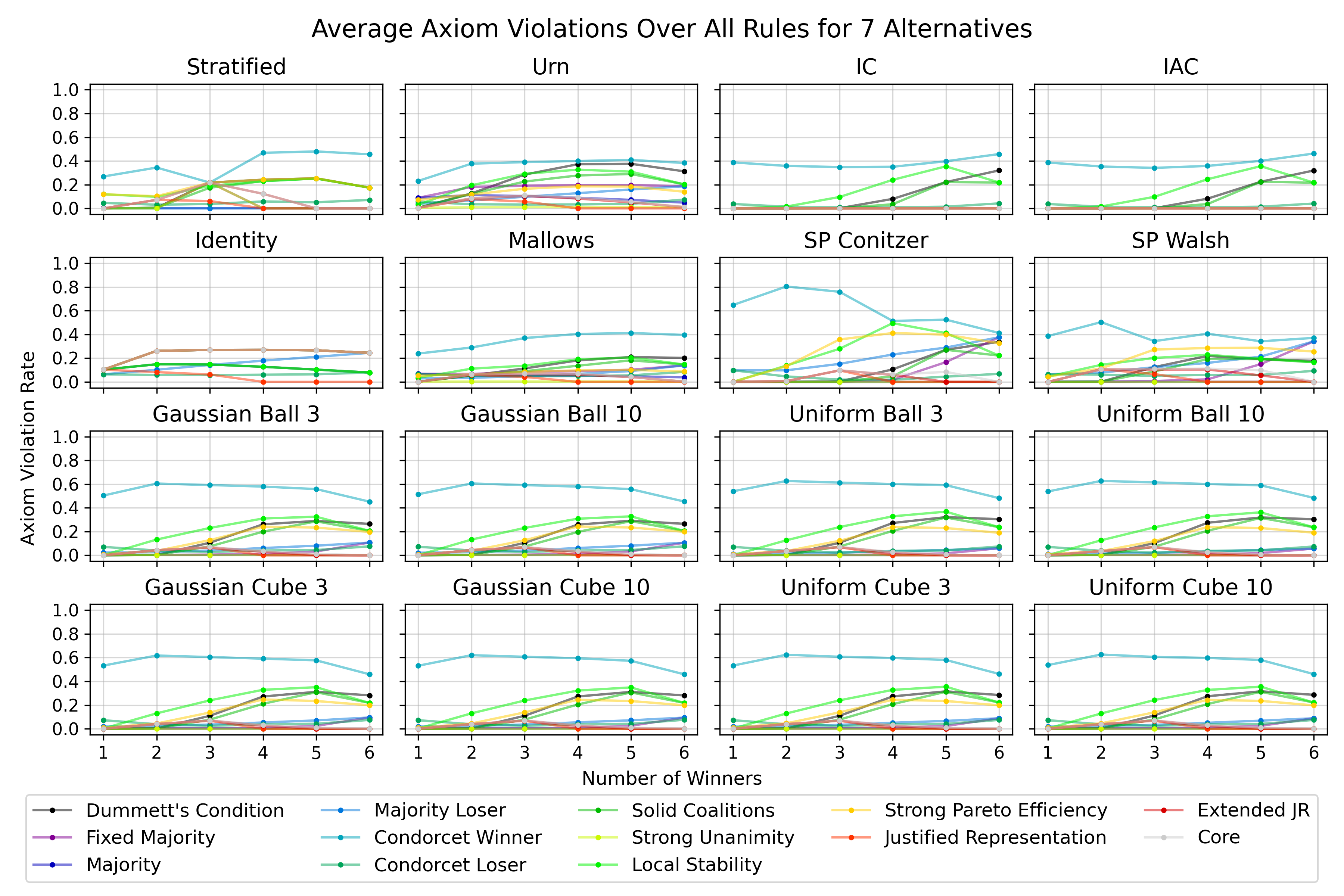}
    \caption{
    Violation rates of each axiom for every preference distribution, averaged over all rules. This shows trends common across all rules in the relationships between axioms and preference distributions.
    }
    \label{fig:violations_by_rule-m=7}
\end{figure}

We visualize the relationship between axioms and preferences in \autoref{fig:violations_by_rule-m=7}. Some previous trends are visible -- the Condorcet Winner axiom is violated frequently in almost all distributions, generally at a similar rate regardless of the number of winners.

\textul{\textbf{Observation:}} Distributions with medium or high levels of structure exhibit quite different axiomatic patterns than other distributions with the same level of structure. This is most surprising in the large differences across Single-Peaked distributions. In fact, on many individual rules Local Stability is violated on Conitzer's Single-Peaked distribution at quite a high rate rate while Walsh's Single-Peaked preferences do not result in violations of Local Stability.

\textul{\textbf{Observation:}} \autoref{fig:violations_by_distribution-m=7} shows a spike in maximum violation rate occurring in most distributions when selecting a committee that contains less than half of alternatives. This is likely an artifact of specific axioms: For example, on Identity preferences it is impossible to violate Justified Representation when choosing a committee containing more than half of alternatives.

\subsubsection{Voting Rules and Preference Distributions}

We now focus primarily on voting rules and preferences. \autoref{fig:violations_by_distribution-m=7} shows the axiom violation rate averaged over all axioms for each different voting rule.
As we would expect, highly structured preference distributions (Identity, Single-Peaked, Stratified) have correspondingly higher axiom violation rates by most rules. This is natural as additional structure provides more opportunity to violate axioms, as shown by \rl{Max} which is much higher for these distributions. Similarly, in distributions with low structure (Impartial Culture, Impartial Anonymous Culture) the maximum violation rate is quite low. Again, we find this intuitive: if all alternatives have equal support then any committee becomes an equally reasonable choice and reasonable axioms should not be violated.

In looking at the behaviour of specific rules on each distribution we can make more fine-grained observations:

\textul{\textbf{Observation:}} On Stratified preferences, $\mathcal{F}^\text{seq-CC}$, $\mathcal{F}^\text{MAV}$ and $\mathcal{F}^\text{CC}$ have higher violation rates than other rules. When $k > 2$, $\mathcal{F}^\text{seq-CC}$ has an AVR very similar to random committees. The ``phase shift'' change in violations as $k > 3$ alternatives are approved correlates with the election of an alternative in the bottom half of the stratified preferences. Recall that there is some set of $\floor{\frac{k}{2}}$ alternatives who are all more preferred by every voter than the other alternatives. When $k > 3$ at least one of these unpopular alternatives is elected which, intuitively, causes violations.

\textul{\textbf{Observation:}} Tie-breaking methods become important when ties are common. While perhaps an obvious statement, this observation explains the unusually high violation rates of \rl{CC}, \rl{seq-CC}, and \rl{SNTV} on Identity preferences.

\subsubsection{Voting Rule Differences}

\begin{table*}[ht]
\centering
\fontsize{7pt}{9pt}\selectfont
\setlength{\tabcolsep}{4.6pt}
\renewcommand{\arraystretch}{1.05}
\begin{tabular}{@{}lcccccccccccccc@{}}
\toprule
 & \rotatebox{90}{Borda} & \rotatebox{90}{EPH} & \rotatebox{90}{SNTV} & \rotatebox{90}{Bloc} & \rotatebox{90}{STV} & \rotatebox{90}{PAV} & \rotatebox{90}{MES} & \rotatebox{90}{CC} & \rotatebox{90}{seq-CC} & \rotatebox{90}{lex-CC} & \rotatebox{90}{Monroe} & \rotatebox{90}{Greedy M.} & \rotatebox{90}{MAV} & \rotatebox{90}{RSD} \\
\midrule
Borda & \cellcolor{blue!0} .000 & -- & -- & -- & -- & -- & -- & -- & -- & -- & -- & -- & -- & -- \\
EPH & \cellcolor{blue!19} .248 & \cellcolor{blue!0} .000 & -- & -- & -- & -- & -- & -- & -- & -- & -- & -- & -- & -- \\
SNTV & \cellcolor{blue!37} .463 & \cellcolor{blue!33} .420 & \cellcolor{blue!0} .000 & -- & -- & -- & -- & -- & -- & -- & -- & -- & -- & -- \\
Bloc & \cellcolor{blue!19} .248 & \cellcolor{blue!1} .021 & \cellcolor{blue!33} .417 & \cellcolor{blue!0} .000 & -- & -- & -- & -- & -- & -- & -- & -- & -- & -- \\
STV & \cellcolor{blue!25} .315 & \cellcolor{blue!28} .360 & \cellcolor{blue!23} .299 & \cellcolor{blue!28} .359 & \cellcolor{blue!0} .000 & -- & -- & -- & -- & -- & -- & -- & -- & -- \\
PAV & \cellcolor{blue!20} .255 & \cellcolor{blue!4} .051 & \cellcolor{blue!34} .429 & \cellcolor{blue!5} .065 & \cellcolor{blue!29} .365 & \cellcolor{blue!0} .000 & -- & -- & -- & -- & -- & -- & -- & -- \\
MES & \cellcolor{blue!21} .271 & \cellcolor{blue!9} .118 & \cellcolor{blue!32} .412 & \cellcolor{blue!10} .130 & \cellcolor{blue!29} .373 & \cellcolor{blue!6} .086 & \cellcolor{blue!0} .000 & -- & -- & -- & -- & -- & -- & -- \\
CC & \cellcolor{blue!48} .600 & \cellcolor{blue!38} .483 & \cellcolor{blue!47} .588 & \cellcolor{blue!38} .487 & \cellcolor{blue!45} .572 & \cellcolor{blue!37} .464 & \cellcolor{blue!39} .496 & \cellcolor{blue!0} .000 & -- & -- & -- & -- & -- & -- \\
seq-CC & \cellcolor{blue!45} .568 & \cellcolor{blue!38} .484 & \cellcolor{blue!37} .464 & \cellcolor{blue!39} .492 & \cellcolor{blue!45} .566 & \cellcolor{blue!37} .467 & \cellcolor{blue!34} .431 & \cellcolor{blue!53} .672 & \cellcolor{blue!0} .000 & -- & -- & -- & -- & -- \\
lex-CC & \cellcolor{blue!24} .310 & \cellcolor{blue!10} .133 & \cellcolor{blue!36} .451 & \cellcolor{blue!11} .144 & \cellcolor{blue!31} .396 & \cellcolor{blue!7} .089 & \cellcolor{blue!9} .117 & \cellcolor{blue!35} .440 & \cellcolor{blue!36} .461 & \cellcolor{blue!0} .000 & -- & -- & -- & -- \\
Monroe & \cellcolor{blue!41} .514 & \cellcolor{blue!31} .396 & \cellcolor{blue!42} .531 & \cellcolor{blue!32} .400 & \cellcolor{blue!39} .491 & \cellcolor{blue!30} .376 & \cellcolor{blue!32} .408 & \cellcolor{blue!9} .117 & \cellcolor{blue!49} .618 & \cellcolor{blue!29} .366 & \cellcolor{blue!0} .000 & -- & -- & -- \\
Greedy M. & \cellcolor{blue!26} .334 & \cellcolor{blue!17} .223 & \cellcolor{blue!34} .428 & \cellcolor{blue!18} .232 & \cellcolor{blue!32} .404 & \cellcolor{blue!16} .203 & \cellcolor{blue!13} .170 & \cellcolor{blue!40} .512 & \cellcolor{blue!32} .402 & \cellcolor{blue!17} .218 & \cellcolor{blue!34} .431 & \cellcolor{blue!0} .000 & -- & -- \\
MAV & \cellcolor{blue!48} .611 & \cellcolor{blue!47} .599 & \cellcolor{blue!55} .694 & \cellcolor{blue!47} .598 & \cellcolor{blue!49} .617 & \cellcolor{blue!47} .597 & \cellcolor{blue!48} .612 & \cellcolor{blue!27} .342 & \cellcolor{blue!65} .813 & \cellcolor{blue!46} .587 & \cellcolor{blue!27} .344 & \cellcolor{blue!50} .635 & \cellcolor{blue!0} .000 & -- \\
RSD & \cellcolor{blue!38} .486 & \cellcolor{blue!37} .465 & \cellcolor{blue!46} .586 & \cellcolor{blue!37} .464 & \cellcolor{blue!42} .526 & \cellcolor{blue!37} .467 & \cellcolor{blue!37} .470 & \cellcolor{blue!51} .646 & \cellcolor{blue!50} .629 & \cellcolor{blue!38} .481 & \cellcolor{blue!46} .577 & \cellcolor{blue!38} .484 & \cellcolor{blue!50} .626 & \cellcolor{blue!0} .000 \\
Random & \cellcolor{blue!57} .714 & \cellcolor{blue!57} .714 & \cellcolor{blue!57} .714 & \cellcolor{blue!57} .714 & \cellcolor{blue!57} .714 & \cellcolor{blue!57} .714 & \cellcolor{blue!57} .714 & \cellcolor{blue!57} .714 & \cellcolor{blue!57} .714 & \cellcolor{blue!57} .714 & \cellcolor{blue!57} .714 & \cellcolor{blue!57} .714 & \cellcolor{blue!57} .714 & \cellcolor{blue!57} .714 \\
Min & \cellcolor{blue!12} .158 & \cellcolor{blue!20} .261 & \cellcolor{blue!38} .476 & \cellcolor{blue!20} .254 & \cellcolor{blue!26} .325 & \cellcolor{blue!22} .278 & \cellcolor{blue!24} .300 & \cellcolor{blue!45} .566 & \cellcolor{blue!47} .598 & \cellcolor{blue!26} .331 & \cellcolor{blue!38} .481 & \cellcolor{blue!28} .361 & \cellcolor{blue!44} .561 & \cellcolor{blue!39} .490 \\
Max & \cellcolor{blue!75} .940 & \cellcolor{blue!74} .936 & \cellcolor{blue!69} .865 & \cellcolor{blue!75} .941 & \cellcolor{blue!72} .912 & \cellcolor{blue!74} .930 & \cellcolor{blue!73} .922 & \cellcolor{blue!65} .813 & \cellcolor{blue!59} .748 & \cellcolor{blue!73} .913 & \cellcolor{blue!69} .866 & \cellcolor{blue!72} .904 & \cellcolor{blue!66} .833 & \cellcolor{blue!68} .856 \\
NN-all & \cellcolor{blue!11} .147 & \cellcolor{blue!19} .240 & \cellcolor{blue!38} .484 & \cellcolor{blue!18} .232 & \cellcolor{blue!26} .335 & \cellcolor{blue!20} .258 & \cellcolor{blue!22} .283 & \cellcolor{blue!45} .569 & \cellcolor{blue!47} .593 & \cellcolor{blue!25} .315 & \cellcolor{blue!38} .484 & \cellcolor{blue!27} .349 & \cellcolor{blue!44} .562 & \cellcolor{blue!38} .484 \\
NN-root & \cellcolor{blue!30} .375 & \cellcolor{blue!31} .395 & \cellcolor{blue!43} .539 & \cellcolor{blue!31} .392 & \cellcolor{blue!33} .424 & \cellcolor{blue!32} .403 & \cellcolor{blue!33} .414 & \cellcolor{blue!42} .530 & \cellcolor{blue!52} .661 & \cellcolor{blue!34} .432 & \cellcolor{blue!32} .411 & \cellcolor{blue!35} .443 & \cellcolor{blue!34} .426 & \cellcolor{blue!39} .496 \\
\bottomrule
\end{tabular}

\caption{Difference between rules for 7 alternatives with $1 \leq k < 7$ averaged over all preference distributions. Darker values correspond to
larger differences. A difference of 0 between two rules indicates the rules always elect the same committee while a difference of 1 indicates
that the rules’ winning committees have maximal overlap.
}
\label{tab:rule_distance_heatmap-m=[7]-pref_dist=all_maintext}
\end{table*}

In \autoref{tab:rule_distance_heatmap-m=[7]-pref_dist=all_maintext} we show the mean difference between each rule based on the overlap between their elected committees.
As expected, randomly chosen committees and $\mathcal{F}^\text{Max}$ are almost always the most different than committees returned by all other rules. Surprisingly, there are exceptions: $\mathcal{F}^\text{MAV}$ and $\mathcal{F}^\text{seq-CC}$ elect committees with  \textit{less} overlap than occurs with a randomly chosen committee. This indicates that these rules optimize for different, and mutually exclusive, goals. Other pairs of rules that are more different from each other than they are random committees can be found in \autoref{app:additional_experiment_results} and our supplementary files, typically involving $\mathcal{F}^\text{MAV}$, $\mathcal{F}^\text{seq-CC}$, and $\mathcal{F}^\text{CC}$. Oddly, $\mathcal{F}^\text{seq-CC}$, and $\mathcal{F}^\text{CC}$ typically elect highly distinct committees despite similar definitions.
Note that all the differences from random committees listed in \autoref{tab:rule_distance_heatmap-m=[7]-pref_dist=all_maintext} are identical, this is merely an effect of rounding. The values are all close, however, due to the nature of random selection of winners.

On the other hand, we can observe rules with quite similar definitions: \rl{Bloc}, \rl{PAV} and \rl{EPH} all follow some procedure that award an equal number of points to a set of alternatives. Each of these rules have very low distance from the others, indicating that the minor differences in their definitions have relatively little effect on the committees they elect.
Unsurprisingly, \rl{NN-all} is most similar to $\mathcal{F}^\text{Min}$. This is understandable given that \rl{NN-all} is trained to find committees that minimize the number of axiom violations. It is also interesting that $\mathcal{F}^\text{Borda}$ is relatively close to $\mathcal{F}^\text{Min}$, though, this is expected given that $\mathcal{F}^\text{Borda}$ was one of the best performing rules in our experiments. Overall, there does seem to be a trend of better performing rules being more similar to one another, and poorer performing rules being more similar to one another.

\subsection{Learning From Learned Rules}

As we see in \autoref{tab:summary_results}, both learning configurations (optimizing for all axioms, and for only root axioms) result in rules with strong axiomatic performance. Averaged across all axioms, \rl{NN-all} has a lower violation rate than any existing rule we evaluated while \rl{NN-root} has a lower violation rate than all rules but \rl{Borda}. While these results are still above the lowest possible violation rate, as given by \rl{Min}, they show that networks are very capable of identifying committees that provide good axiomatic properties.

By learning two different sets of axioms we are able to both deepen our understanding of which axioms are more difficult to learn and to gauge whether learning ``redundant'' axioms with \rl{NN-all} is beneficial. As \rl{NN-root} has a much lower violation rate, it is clearly beneficial to learn from all axioms. This is intuitive, \rl{NN-all} has the opportunity to learn to satisfy non-root axioms even in cases when root axioms may not be possible to mutually satisfy.
However, even when evaluating \textit{only the root axioms}, \rl{NN-all} has an average violation rate of $0.0283$ compared to an average violation rate of $0.0692$ for \rl{NN-root}. Training on the additional axioms appears to provide enough additional signal to the learning process which reduces violations of root axioms.

\subsection{Real-World Data}

Using networks trained on the Mixed distribution, we applied each \rl{NN-all} to a selection of real-world data collected from PrefLib ~\cite{mattei2013preflib}. 
We find that this highlights the sensitivity of learned models to their training data: our Mixed distribution is not a perfect match of the distribution underlying the empirical data we tested on. Nonetheless, we find that \rl{NN-all} generalizes well, having, for example, a mean AVR of $0.102$ compared to an AVR of $0.303$ for committees selected at random. For more details and results of these experiments we refer to \autoref{app:real_world}.

\subsection{Optimized Positional Scoring Rules}

The low axiom violation rate of \rl{Borda} inspires a natural follow-up question: Is there a positional scoring rule which provides even stronger performance than \rl{Borda}? A positive answer here would be interesting both to practitioners and theorists. The black box nature of the neural networks underlying each \rl{NN} restrict these rules from being applied in settings where interpretability and reliability are vital. On the other hand, positional scoring rules provide an immediate intuition as to the behaviour of the rule that allows them to be applied in practice (e.g., the Plurality rule can be said to reward voter's favourite alternatives while the Anti-Plurality rule punishes those alternatives least preferred by voters ~\cite{brandt2016handbook}). Theoreticians may find novel positional scoring rules interesting as such rules could point in the direction of new results or points of analysis.

\begin{wraptable}{r}{7.4cm}
\centering
\begin{tabular}{@{}ccccc@{}}
    & \multicolumn{2}{c}{\textbf{All Axioms}} & \multicolumn{2}{c}{\textbf{Root Axioms}} \\ \midrule
$k$ & \rl{Borda} & \rl{Opt}                   & \rl{Borda}           & \rl{Opt}          \\ \midrule
1   & 0.010      & \multicolumn{1}{c|}{0.008} & 0.019                & 0.016             \\
2   & 0.011      & \multicolumn{1}{c|}{0.011} & 0.021                & 0.018             \\
3   & 0.022      & \multicolumn{1}{c|}{0.018} & 0.036                & 0.032             \\
4   & 0.030      & \multicolumn{1}{c|}{0.025} & 0.044                & 0.043             \\
5   & 0.034      & \multicolumn{1}{c|}{0.028} & 0.042                & 0.044             \\
6   & 0.025      & \multicolumn{1}{c|}{0.022} & 0.030                & 0.027             \\ \bottomrule
\end{tabular}
\caption{Axiom violation rate for $m=7$ alternatives on both axiom sets by \rl{Borda} and the positional scoring rule found via simulated annealing, \rl{Opt}.}
\label{tab:annealed-psr}
\end{wraptable}

As positional scoring rules have a natural ``state'' (their score vector) we are easily able to apply generic optimization techniques to find new vectors which may have low axiom violation rates. To do this, we use the \texttt{optimal-voting} package \cite{armstrong2025optimal} which uses simulated annealing to optimize a score vector. We set the loss function equal to the axiom violation rate and run annealing for Mixed preferences using $m=7$ alternatives only, with one annealing run for each number of winners. We refer to a rule optimized in this manner as \rl{Opt}. Due to computational and time constraints we run for only 1000 steps, using 2000 profiles sampled from the larger set of 25,000 profiles. All rules have an initial state corresponding to the Borda rule. We then evaluate the violation rates of each rule on a different set of 25,000 profiles from the Mixed distribution to ensure that our results do not simply reflect overfitting on the training data.

\autoref{tab:annealed-psr} shows the results of annealing using all axioms in the loss function, and using just root axioms. The vectors resulting from annealing are included in \Cref{app:optimized_psr}. We highlight that our optimization process is quite limited in both data and number of steps. Despite this, optimization consistently finds some vector that outperforms \rl{Borda}. These vectors also display a common trend; each optimized score vector awards higher points to the first $k$ alternatives, then significantly reduces the number of points awarded.
This suggests a trend towards a vector similar to $k$-approval (where each voter's most preferred $k$ alternatives receive a single point). However, we note that the $k$-approval rule itself proves to have far worse axiomatic properties than both \rl{Opt} and \rl{Borda}. These preliminary results leave open the possibility of some novel class of score vector which follows a \textit{different} decreasing sequence than \rl{Borda}.

\section{Discussion}

We have shown that (1) different multi-winner voting rules elect committees that are distinct from one another and  violate axioms at  different rates, (2) these differences depend greatly upon the underlying preferences of voters, and (3) it is possible to learn novel rules which result in significantly lower axiom violation rates. These contributions join a growing body of work demonstrating the effectiveness of combining machine learning with social choice \cite{golowich2018deep,conitzer2024social,lanctot2025soft} and adds to the literature extending axiomatic analysis beyond the worst case \cite{flanigan2023smoothed}. We highlight findings of particular note to the broader research community along with and opportunities for future work.

\noindent
\textbf{Competing Definitions of Proportionality}
We observe two contrasting types of proportionality among our axioms. Core, EJR, and JR are formally linked (each a weaker version of the former) \cite{lackner2023multi} while Local Stability, Dummett's, and Solid Coalitions have distinct origins. Most voting rules violate the first three of these axioms similarly, while there is much more variability in AVR on axioms from the second type.
Clear distinctions between these two types of proportionality are missing in the literature.
For example, $\mathcal{F}^\text{CC}$ and $\mathcal{F}^\text{Monroe}$ ``explicitly aim at proportional representation'' \cite{elkind2017properties} yet these rules violate Dummett's Condition, Solid Coalitions, and Local Stability frequently while maintaining a low AVR on JR, EJR, and the Core.

\noindent
\textbf{Proportionality and Electing Losers} 
The high AVR of the proportional rules $\mathcal{F}^\text{CC}$ and $\mathcal{F}^\text{Monroe}$ for the Condorcet Loser and Majority Loser axioms demonstrates a fundamental tension between (some types of) proportionality and avoiding bad alternatives. To satisfy proportionality, it may be necessary to elect an alternative ranked low by a majority of voters. New axioms may aim to bridge this gap by formally considering a balance between proportional representation and electing highly polarizing alternatives.

\noindent
\textbf{Individual Excellence and Proportional Rules} Rules focused primarily on electing alternatives which are individually popular among voters generally result in fewer violations of axioms focused on proportionality than rules with a state goal of proportionality. If proportionality guarantees can be established for individual excellence rules this would allow benefitting from their superior performance in settings where proportionality is important.

\noindent
\textbf{Learning Simple Axioms} Targetting both simple and complex axioms resulted in \rl{NN-all} having superior performance to \rl{NN-root}. This suggests that future axiom learning tasks could be enhanced by introducing simple dummy axioms that provide some feedback about more complex axioms (e.g., a majority winner must always be in a Condorcet winning set, so the Majority Winner axiom may be seen as informative about Condorcet Winner).

\noindent
\textbf{Difference Between Rules}
Our results can be qualitatively compared with existing maps of multi-winner elections \cite{faliszewski2023experimental}. In particular, in both their work and ours: (1) rules focused on individual excellence elect different committees than proportional rules, and (2) $\mathcal{F}^\text{MAV}$ is very different from all other rules.

\subsection{Future Directions}
There are several interesting directions that this work can take. First, our research can be directly extended to support arbitrary sets of intraprofile axioms and extensions to the more general class of \textit{interprofile} axioms \cite{schmidtlein2022voting,schmidtlein2023voting}.
This would allow us to measure violations of additional axioms such as consistency or clone-proofness \cite{brandt2016handbook}.

Second, we proposed a new metric for evaluating voting rules, the axiom violation rate. While impossibility theorems show that some axioms can never be mutually satisfied; in these cases, an interesting challenge for future research is to develop new rules optimized for low axiom violation rate. Using black-box machine learning techniques may be sufficient, however identifying classes of rule (such as positional scoring rules) which can be optimized in a way that provides a reliable, interpretable outcome will provide additional usability.

Finally, we are interested in extending our metric of axiom violations to allow robust, theoretical conclusions about axiomatic properties -- similar to those provided by the PAC-learning framework \cite{valiant1984theory}. We believe that such an approach would complement both the axiomatic and the empirical studies of voting rules.

\bibliography{refs}

\appendix

\section{Details on Axiom Relationships}
\label{app:axiom_implications}

\begin{figure}[t]
\centering
\begin{tikzpicture}[
    every node/.style={draw, ellipse, fill=blue!10, text=black, font=\small, minimum width=2.5cm, minimum height=1cm, align=center},
    arr/.style={-stealth, thick},
    incomplete/.style={-stealth, thick, red},
    dashedarr/.style={dashed, -stealth, thick},
    incompletedashed/.style={dashed, -stealth, thick, red},
    labelstyle/.style={font=\tiny, black, fill=blue!30}, 
    greennode/.style={fill=green!20}, 
    label/.style={fill=white, font=\tiny, inner sep=2pt}
]

    \node (CW) at (-4.5, -2) {Condorcet \\ Winner};
    \node (FM) at (-3, 0.5) {Fixed \\ Majority};
    \node (MW) at (-4, 6) {Majority \\ Winner}; 

    \node (SC) at (0, 6) {Solid \\ Coalitions};
    \node[greennode] (JR) at (5, 6) {Justified \\ Representation};
    \node[greennode] (EJR) at (5, 3) {EJR};
    \node[greennode] (Core) at (5, 0) {Core};

    \node (D) at (2, 3) {Dummett's}; 
    \node (LS) at (-1.5, 3) {Local\\ Stability
    }; 
    \node (SU) at (0.5, 0.5) {Strong \\ Unanimity};
    \node[greennode] (SP) at (3, -2) {Strong \\ Pareto};

    \node (CL) at (-6, 4) {Condorcet \\ Loser};
    \node (ML) at (-6, 1.5) {Majority \\ Loser};

    \draw[arr] (CW) -- (FM);
    \draw[arr] (FM) -- (SU);

    \draw[arr] (Core) -- (EJR);
    \draw[arr] (EJR) -- (JR);
    
    \draw[dashedarr] (SC) -- (JR);

    \draw[arr] (SC) -- (MW) node[midway, above, xshift=0mm, draw=none, fill=none, font=\tiny, inner sep=2pt, minimum width=0pt, minimum height=0pt] {$k \geq 2$};
    
    \draw[arr] (LS) -- (SC);
    \draw[arr] (D) -- (SC);
    \draw[arr] (LS) -- (SU);

    \draw[arr] (LS) -- (CL) node[midway, above, xshift=0mm, draw=none, fill=none, font=\tiny, inner sep=2pt, minimum width=0pt, minimum height=0pt] {$k \geq 2$};

    \draw[dashedarr] (SP) -- (SU);
    \draw[dashedarr] (EJR) -- (SU);
    \draw[arr] (D) -- (SU);

\end{tikzpicture}
\caption{
Satisfaction relationships between axioms that we consider. An arrow from node X to node Y indicates that satisfying axiom X also leads to satisfaction of axiom Y, or the equivalent contrapositive: violating Y also implies a violation of X. Blue nodes are those originating in ordinal settings while green nodes originate in approval-based settings. A dashed line indicates that the connection relies upon our assumption that each voter approves of their $k$ top-ranked alternatives.
\\
\textbf{Root Nodes:} Local Stability, Dummett's, Condorcet Winner, Strong Pareto, Core, Majority Loser.
}
\label{fig:axiom_implications}
\end{figure}
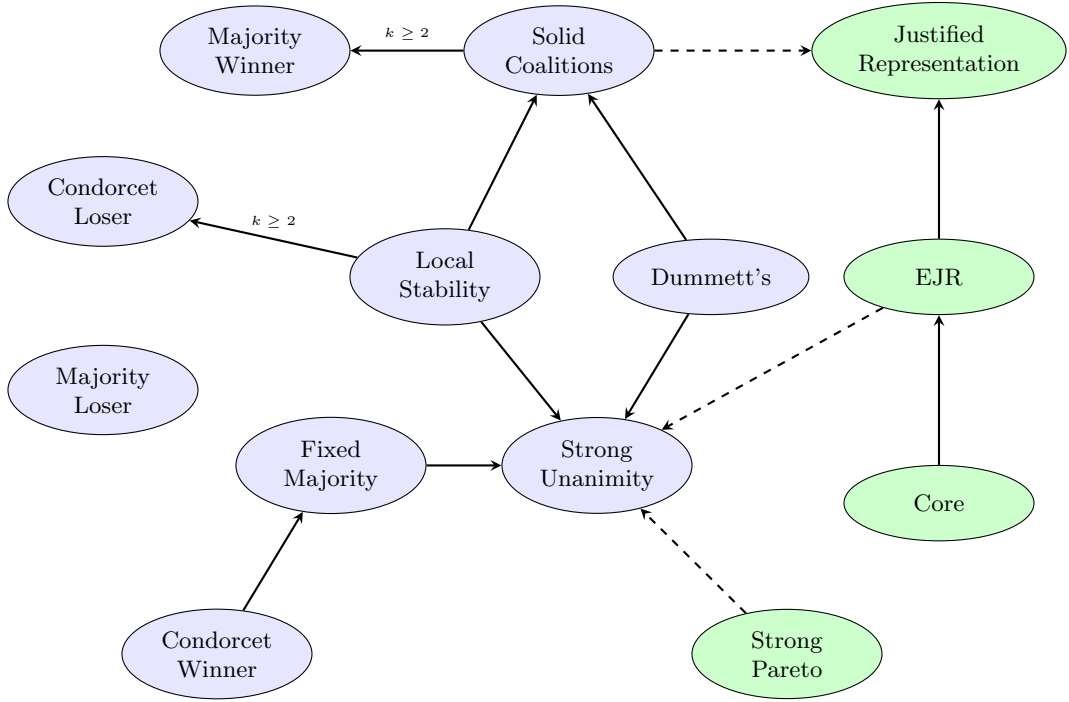

This section describes the satisfaction relationships between the axioms we consider. We show in \autoref{fig:axiom_implications} that many of the axioms we use are necessarily satisfied if some other axiom is satisfied. Some of these relationships have been previously described while several represent novel connections. We do not claim that these results are complete; other connections may exist which we have not identified. This section serves to provide some intuition as to which axioms are more or less similar to others.

\subsection{Known Relationships}

We first list pairs of axioms for which we know that one implies the other.
As this is not the primary focus of our paper, our proofs remain at a high level and are relatively informal.

\begin{relation}
   Condorcet Winner $\implies$ Fixed Majority 
\end{relation}

A Fixed Majority is a set of voters that is a Condorcet winning set. Thus, satisfying the Condorcet Winner property  will also lead to satisfaction of Fixed Majority.


\begin{relation}
   If the selected committee has at least $k\geq 2$ alternatives, then Solid Coalitions $\implies$ Majority Winner  
\end{relation}


Let $X$ refer to the set of all alternatives ranked first by at least $\frac{n}{k}$ voters. If Solid Coalitions is satisfied then every member of $X$ is in the winning committee. A majority winner is one ranked first by $\frac{n}{2} \geq \frac{n}{k}$ voters. Thus, if Solid Coalitions is satisfied and a majority winner exists, then it is elected.
Note that this is not the case when $k=1$. Then, $X$ contains only an alternative that is unanimously ranked first which may not include a majority winner.


\begin{relation} If every voter approves of exactly their $k$ top-ranked alternatives, then 
Solid Coalitions $\implies$ Justified Representation.    
\end{relation}

In order to satisfy JR, whenever at least $\frac{n}{k}$ voters approve of an alternative, one of those voters must approve of one of the elected alternatives.
SC requires that whenever $\frac{n}{k}$ voters rank some alternative $c$ first, this alternative is in the winning set. 

If SC is satisfied then, by assumption, whenever there exists a set of at least $\frac{n}{k}$ voters who all rank an alternative first, they also approve that alternative and elect it.



\begin{relation}
If a locally stable committee exists, then Local Stability $\implies$ Solid Coalitions.
\end{relation}

Consider a locally stable committee $C$. Say there exists a set of voters $V^*$ with $|V^*| \geq \frac{n}{k}$ who all rank candidate $x$ first. If $x \in C$ then Solid Coalitions is satisfied.
If $x \notin C$ then the committee is not locally stable and we have a contradiction. Thus, a locally stable committee will always elect an alternative required by Solid Coalitions.



\begin{relation}
    If a locally stable committee exists, then Local Stability $\implies$ Strong Unanimity.
\end{relation}

Consider a locally stable committee $C$ and a profile in which all voters rank the same $k$ alternatives above all others. If $C$ were not exactly the $k$ top-ranked alternatives, then there would be some alternative not in $C$ preferred by all voters, and $C$ would not be locally stable.



\begin{relation}
 If every voter approves of exactly their $k$ top-ranked alternatives, then  EJR $\implies$ Strong Unanimity.  
\end{relation}
If there exists a profile in which all voters rank the same $k$ alternatives above all others (and, by assumption, are exactly the approval set of each voter), these voters form a $k$-cohesive group and, if EJR is satisfied, will form the winning committee.



\begin{relation}
    If every voter approves of exactly their $k$ top-ranked alternatives, then Strong Pareto $\implies$ Strong Unanimity.
\end{relation}

Say there exists a profile in which all voters rank the same $k$ alternatives above all others and, by assumption, form the approval set of each voter.
Any committee which is not exactly these $k$ alternatives will be dominated by a committee containing these $k$ alternatives so when voters are unanimous Strong Pareto is sufficient to elect their preferred committee.



\begin{relation}
   If a locally stable committee exists with at least $k\geq 2$ members, then Local Stability $\implies$ Condorcet Loser. 
\end{relation}
For any preference profile with a Condorcet losing committee $C$, there exists some alternative $x \notin C$ which is preferred by at least $\frac{n}{2}$ voters to every alternative in $C$.
Under our assumption that $k \geq 2$, a locally stable committee cannot contain any alternative $y$ such that $x$ is preferred to $y$ by more than $\frac{n}{2}$ voters. Thus, a locally stable committee is not a Condorcet Loser.

\begin{relation}
Fixed Majority $\implies$ Strong Unanimity.
\end{relation}
If every voter ranks the same $k$ alternatives on top, those alternatives constitute a fixed majority committee and will be elected if FM is satisfied.

\begin{relation}
    Dummett's $\implies$ Solid Coalitions.
\end{relation}

We refer to \citet{elkind2017properties} for discussion of this relationship.

\begin{relation}
Dummett's $\implies$ Unanimity.
\end{relation}

We refer to \citet{elkind2017properties} for discussion of this relationship.

\begin{relation}
    Core $\implies$ EJR $\implies$ JR.
\end{relation}

We refer to \citet{lackner2023multi} for discussion of these relationships.






\subsection{Relationships Known To Not Exist}

We also include some counter-examples showing certain relationships that \textit{do not} exist. While it is the case that \autoref{fig:axiom_implications} may be missing edges we show here some of the edges that we thought may exist due to the similar nature of certain axioms, but can be demonstrated not to exist.

\begin{relation}
Condorcet Winner $\notimplies$ Dummett's and Dummett's $\notimplies$ Condorcet Winner.
\end{relation}


Say that $n = 10$, $l = 2$, $k = 5$  and consider voters with the following preference orders:

{
\centering
\begin{tabular}{cc}
    $v_0$ &  $0 \succ 1 \succ 2 \succ 3 \succ 4 \succ 5 \succ 6 \succ 7 \succ 8 \succ 9$ \\
    $v_1$ & $0 \succ 1 \succ 4 \succ 5 \succ 6 \succ 7 \succ 8 \succ 9 \succ 2 \succ 3$ \\
    $v_2$ & $0 \succ 1 \succ 6 \succ 7 \succ 8 \succ 9 \succ 2 \succ 3 \succ 4 \succ 5$ \\
    $v_3$ & $0 \succ 1 \succ 8 \succ 9 \succ 2 \succ 3 \succ 4 \succ 5 \succ 6 \succ 7$ \\
    $6 \times v_\text{maj}$ & $9 \succ 8 \succ 7 \succ 6 \succ 5 \succ 4 \succ 3 \succ 2 \succ 1 \succ 0$ \\
\end{tabular}
}

A group of $\frac{l \cdot n}{k} = 4$ voters ($v_0, v_1, v_2, v_3$) rank the same 2 alternatives, $\{0, 1\}$, on top. These alternatives must be in the winning committee in order to satisfy Dummett's Condition.
However, the Condorcet winning set is $\{9, 8, 7, 6, 5\}$. 
In this case, any possible winning committee will violate at least one of Condorcet Winner and Dummett's condition.

\begin{relation}
EJR $\notimplies$ Dummett's.
\end{relation}


Consider the following preference profile and say we are choosing $k = 5$ winners. For convenience, a vertical bar separates the approved alternatives and the disapproved alternatives.

{
\centering
\begin{tabular}{cc}
    $v_0$ &  $0 \succ 6 \succ 2 \succ 3 \succ 4 | \succ 5 \succ 1 \succ 7 \succ 8 \succ 9$ \\
    $v_1$ & $0 \succ 7 \succ 2 \succ 3 \succ 4 | \succ 5 \succ 6 \succ 1 \succ 8 \succ 9$ \\
    $v_2$ & $0 \succ 8 \succ 2 \succ 3 \succ 4 | \succ 5 \succ 6 \succ 7 \succ 1 \succ 9$ \\
    $v_3$ & $0 \succ 9 \succ 2 \succ 3 \succ 4 | \succ 5 \succ 6 \succ 7 \succ 8 \succ 1$ \\
    $6 \times v_\text{maj}$ & $9 \succ 8 \succ 7 \succ 6 \succ 5 | \succ 4 \succ 3 \succ 2 \succ 1 \succ 0$ \\
\end{tabular}
}

In the profile above, note that $v_0, v_1, v_2, v_3$ form a 2-cohesive group. In order to satisfy, EJR one of these voters must approve of two elected alternatives.
Similarly, the other 6 voters form a 3-cohesive group and must approve of three elected alternatives.
A committee which would not violate EJR is ${3, 4, 5, 6, 7}$

Dummett's condition requires that, since 4 voters rank $0$ on top, $0$ must be in any winning committee. Thus, it is possible to violate Dummett's condition without violating EJR. Satisfaction of EJR does not imply satisfaction of Dummett's condition.

\begin{relation}
    Dummett's $\notimplies$ EJR.
\end{relation}




{
\centering
\begin{tabular}{cc}
    $v_0$ &  $0 \succ 6 \succ 2 \succ 3 \succ 4 | \succ 5 \succ 1 \succ 7 \succ 8 \succ 9$ \\
    $v_1$ & $1 \succ 7 \succ 2 \succ 3 \succ 4 | \succ 5 \succ 6 \succ 0 \succ 8 \succ 9$ \\
    $v_2$ & $2 \succ 8 \succ 1 \succ 3 \succ 4 | \succ 5 \succ 6 \succ 7 \succ 0 \succ 9$ \\
    $v_3$ & $3 \succ 9 \succ 2 \succ 1 \succ 4 | \succ 5 \succ 6 \succ 7 \succ 8 \succ 0$ \\
    $6 \times v_\text{maj}$ & $9 \succ 8 \succ 7 \succ 6 \succ 5 | \succ 4 \succ 3 \succ 2 \succ 1 \succ 0$ \\
\end{tabular}
}

Consider the above profile. Dummett's condition requires that the top 3 preferences of the group of 6 voters must be elected, but does not capture any other group information.
A committee elected by a rule satisfying Dummett's condition may be $\{9, 8, 7, 6, 5\}$.

However, $v_0, v_1, v_2, v_3$ still form a $2$-cohesive group -- each of the voters approves of $\{3, 4\}$. In order to satisfy EJR, one of these voters must approve of two elected alternatives which is not the case for the winning committee $\{9, 8, 7, 6, 5\}$. Thus, satisfying Dummett's does not imply satisfaction of EJR.

\begin{relation}
    Dummett's $\notimplies$ Fixed Majority and Fixed Majority $\notimplies$ Dummett's.
\end{relation}
Say that $n = 10$, $l = 2$, $k = 5$  and consider voters with the following preference orders:

{
\centering
\begin{tabular}{cc}
    $v_0$ &  $0 \succ 1 \succ 2 \succ 3 \succ 4 \succ 5 \succ 6 \succ 7 \succ 8 \succ 9$ \\
    $v_1$ & $0 \succ 1 \succ 4 \succ 5 \succ 6 \succ 7 \succ 8 \succ 9 \succ 2 \succ 3$ \\
    $v_2$ & $0 \succ 1 \succ 6 \succ 7 \succ 8 \succ 9 \succ 2 \succ 3 \succ 4 \succ 5$ \\
    $v_3$ & $0 \succ 1 \succ 8 \succ 9 \succ 2 \succ 3 \succ 4 \succ 5 \succ 6 \succ 7$ \\
    $6 \times v_\text{maj}$ & $9 \succ 8 \succ 7 \succ 6 \succ 5 \succ 4 \succ 3 \succ 2 \succ 1 \succ 0$ \\
\end{tabular}
}

As in previous example (D and CW), consider the required winners from each axiom.

Dummett's requires that -- since $\frac{ln}{k} = \frac{2 \cdot 10}{5} = 4$ voters rank 0 and 1 first -- 0 and 1 must be winners.
There is a Fixed Majority which all rank candidates 5 through 9 first. To satisfy FM, these must be the winners.
In this case it is possible to satisfy one of Dummett's condition or Fixed Majority, but not both.

    




\clearpage
\section{Results on Real World Data}
\label{app:real_world}

We gathered from PrefLib (an online repository of real-world preference information) all complete preference profiles containing 5, 6, or 7 alternatives \cite{mattei2013preflib}. Resulting in, respectively, 15, 19, and 22 preference profiles. For each of these preferences, we identified through brute force search the committees that would violates the fewest and most of our axioms. 

As this amount of data is not suitable for a meaningful amount of learning, we measured the Axiom Violation Rate on this data of networks trained on our Mixed preference distributions. Results of this evaluation are found in the next section. These results emphasize clearly that the Mixed preference distribution is mildly informative about the distribution(s) underlying PrefLib data but that there is also a significant gap between the training and testing distributions in this area. That is, \rl{NN-all} has worse AVR than some rules but also much better AVR than many rules. Complete results of the PrefLib evaluation for each number of alternatives are listed below.

\begin{itemize}
\item \nameref{sec:5_alternatives-Real World Data_preferences}
\item \nameref{sec:6_alternatives-Real World Data_preferences}
\item \nameref{sec:7_alternatives-Real World Data_preferences}
\end{itemize}

\subsection{5 Alternatives, Real World Data}
\label{sec:5_alternatives-Real World Data_preferences}
\begin{table}[ht]
\label{tab:summary_table-n_profiles=[25000]-num_voters=[50]-m=[5]-pref_dist=['rwd']-axioms=['all']}
\centering
\fontsize{7pt}{9pt}
\selectfont
\setlength{\tabcolsep}{4.6pt}
\renewcommand{\arraystretch}{1.05}\begin{tabular}{lc|cccccc|ccccccc}
\toprule
Method & \rotatebox{90}{Mean} & \rotatebox{90}{Maj W} & \rotatebox{90}{\underline{Maj L}} & \rotatebox{90}{\underline{Cond W}} & \rotatebox{90}{Cond L} & \rotatebox{90}{\underline{Pareto}} & \rotatebox{90}{F Maj} & \rotatebox{90}{Unanimity} & \rotatebox{90}{\underline{Dummett's}} & \rotatebox{90}{JR} & \rotatebox{90}{EJR} & \rotatebox{90}{\underline{Core}} & \rotatebox{90}{S. Coalitions} & \rotatebox{90}{\underline{Stability}} \\
\midrule
NN-all & .081 & .071 & .018 & .428 & .018 & .112 & .089 & \textbf{0} & .154 & \textbf{0} & \textbf{0} & \textbf{0} & .071 & .089 \\
Min & \textbf{.004} & \textbf{0} & \textbf{0} & \textbf{0} & \textbf{0} & .018 & \textbf{0} & \textbf{0} & .036 & \textbf{0} & \textbf{0} & \textbf{0} & \textbf{0} & \textbf{0} \\
Max & .471 & .286 & .357 & .929 & .554 & .679 & .304 & \textbf{0} & .661 & .214 & .357 & .357 & .625 & .804 \\
\midrule
Borda & .015 & \textbf{0} & \textbf{0} & .054 & \textbf{0} & .018 & \textbf{0} & \cellcolor{green!25}\textbf{0} & .054 & \textbf{0} & \textbf{0} & \textbf{0} & .036 & .036 \\
EPH & .038 & \textbf{0} & \textbf{0} & .107 & \textbf{0} & \cellcolor{green!25}\textbf{0} & \textbf{0} & \textbf{0} & .179 & \textbf{0} & \textbf{0} & \textbf{0} & .107 & .107 \\
SNTV & .055 & \cellcolor{green!25}\textbf{0} & .089 & .339 & \textbf{0} & .125 & .125 & \textbf{0} & .036 & \textbf{0} & \textbf{0} & \textbf{0} & \cellcolor{green!25}\textbf{0} & \textbf{0} \\
Bloc & .038 & \textbf{0} & \textbf{0} & .107 & \textbf{0} & \cellcolor{green!25}\textbf{0} & \cellcolor{green!25}\textbf{0} & \cellcolor{green!25}\textbf{0} & .179 & \textbf{0} & \textbf{0} & \textbf{0} & .107 & .107 \\
\midrule
STV & .019 & \cellcolor{green!25}\textbf{0} & .018 & .161 & \textbf{0} & .036 & .036 & \textbf{0} & \textbf{0} & \textbf{0} & \textbf{0} & \textbf{0} & \cellcolor{green!25}\textbf{0} & \textbf{0} \\
PAV & .044 & \textbf{0} & \textbf{0} & .143 & \textbf{0} & \cellcolor{green!25}\textbf{0} & \textbf{0} & \textbf{0} & .179 & \cellcolor{green!25}\textbf{0} & \cellcolor{green!25}\textbf{0} & \textbf{0} & .125 & .125 \\
MES & .045 & \textbf{0} & \textbf{0} & .161 & \textbf{0} & \textbf{0} & \textbf{0} & \textbf{0} & .179 & \cellcolor{green!25}\textbf{0} & \cellcolor{green!25}\textbf{0} & \textbf{0} & .125 & .125 \\
CC & .136 & .071 & .125 & .446 & .018 & .161 & .196 & \textbf{0} & .304 & \cellcolor{green!25}\textbf{0} & \textbf{0} & \textbf{0} & .214 & .232 \\
seq-CC & .146 & .018 & .125 & .554 & .018 & .232 & .196 & \textbf{0} & .339 & \cellcolor{green!25}\textbf{0} & \textbf{0} & \textbf{0} & .196 & .214 \\
lex-CC & .063 & \textbf{0} & .036 & .196 & \textbf{0} & \textbf{0} & .054 & \textbf{0} & .214 & \textbf{0} & \textbf{0} & \textbf{0} & .161 & .161 \\
Monroe & .077 & .036 & .054 & .286 & .018 & .071 & .071 & \cellcolor{green!25}\textbf{0} & .196 & \cellcolor{green!25}\textbf{0} & \textbf{0} & \textbf{0} & .125 & .143 \\
Greedy M. & .059 & \textbf{0} & .018 & .232 & \textbf{0} & .018 & .036 & \cellcolor{green!25}\textbf{0} & .196 & \cellcolor{green!25}\textbf{0} & \textbf{0} & \textbf{0} & .125 & .143 \\
\midrule
MAV & .162 & .125 & .143 & .607 & .054 & .179 & .250 & \textbf{0} & .268 & \textbf{0} & \textbf{0} & \textbf{0} & .196 & .286 \\
RSD & .104 & .089 & .036 & .321 & .018 & \textbf{0} & .107 & \textbf{0} & .214 & .054 & .054 & .054 & .161 & .250 \\
Random & .255 & .196 & .214 & .732 & .107 & .357 & .286 & \textbf{0} & .464 & .071 & .089 & .089 & .304 & .411 \\
\bottomrule
\end{tabular}
\caption{Average Axiom Violation Rate for 5 alternatives and $1 \leq k < 5$ winners across Real World Data preferences.}
\end{table}

\begin{table}[ht]
\centering
\fontsize{7pt}{9pt}\selectfont
\setlength{\tabcolsep}{4.6pt}
\renewcommand{\arraystretch}{1.05}
\begin{tabular}{@{}lcccccccccccccc@{}}
\toprule
 & \rotatebox{90}{Borda} & \rotatebox{90}{EPH} & \rotatebox{90}{SNTV} & \rotatebox{90}{Bloc} & \rotatebox{90}{STV} & \rotatebox{90}{PAV} & \rotatebox{90}{MES} & \rotatebox{90}{CC} & \rotatebox{90}{seq-CC} & \rotatebox{90}{lex-CC} & \rotatebox{90}{Monroe} & \rotatebox{90}{Greedy M.} & \rotatebox{90}{MAV} & \rotatebox{90}{RSD} \\
\midrule
Borda & 0 & -- & -- & -- & -- & -- & -- & -- & -- & -- & -- & -- & -- & -- \\
EPH & \cellcolor{blue!17} .223 & 0 & -- & -- & -- & -- & -- & -- & -- & -- & -- & -- & -- & -- \\
SNTV & \cellcolor{blue!24} .304 & \cellcolor{blue!30} .375 & 0 & -- & -- & -- & -- & -- & -- & -- & -- & -- & -- & -- \\
Bloc & \cellcolor{blue!17} .223 & 0 & \cellcolor{blue!30} .375 & 0 & -- & -- & -- & -- & -- & -- & -- & -- & -- & -- \\
STV & \cellcolor{blue!14} .187 & \cellcolor{blue!24} .304 & \cellcolor{blue!19} .241 & \cellcolor{blue!24} .304 & 0 & -- & -- & -- & -- & -- & -- & -- & -- & -- \\
PAV & \cellcolor{blue!17} .214 & \cellcolor{blue!2} .036 & \cellcolor{blue!30} .384 & \cellcolor{blue!2} .036 & \cellcolor{blue!25} .321 & 0 & -- & -- & -- & -- & -- & -- & -- & -- \\
MES & \cellcolor{blue!15} .196 & \cellcolor{blue!9} .116 & \cellcolor{blue!26} .330 & \cellcolor{blue!9} .116 & \cellcolor{blue!24} .304 & \cellcolor{blue!7} .089 & 0 & -- & -- & -- & -- & -- & -- & -- \\
CC & \cellcolor{blue!32} .411 & \cellcolor{blue!24} .304 & \cellcolor{blue!38} .482 & \cellcolor{blue!24} .304 & \cellcolor{blue!35} .446 & \cellcolor{blue!24} .304 & \cellcolor{blue!31} .393 & 0 & -- & -- & -- & -- & -- & -- \\
seq-CC & \cellcolor{blue!36} .455 & \cellcolor{blue!39} .491 & \cellcolor{blue!27} .348 & \cellcolor{blue!39} .491 & \cellcolor{blue!39} .491 & \cellcolor{blue!37} .473 & \cellcolor{blue!32} .402 & \cellcolor{blue!49} .616 & 0 & -- & -- & -- & -- & -- \\
lex-CC & \cellcolor{blue!19} .241 & \cellcolor{blue!5} .071 & \cellcolor{blue!31} .393 & \cellcolor{blue!5} .071 & \cellcolor{blue!26} .330 & \cellcolor{blue!2} .036 & \cellcolor{blue!10} .125 & \cellcolor{blue!22} .277 & \cellcolor{blue!39} .491 & 0 & -- & -- & -- & -- \\
Monroe & \cellcolor{blue!26} .330 & \cellcolor{blue!16} .205 & \cellcolor{blue!31} .393 & \cellcolor{blue!16} .205 & \cellcolor{blue!27} .339 & \cellcolor{blue!16} .205 & \cellcolor{blue!23} .295 & \cellcolor{blue!8} .107 & \cellcolor{blue!43} .545 & \cellcolor{blue!16} .205 & 0 & -- & -- & -- \\
Greedy M. & \cellcolor{blue!21} .268 & \cellcolor{blue!17} .214 & \cellcolor{blue!24} .304 & \cellcolor{blue!17} .214 & \cellcolor{blue!26} .330 & \cellcolor{blue!16} .205 & \cellcolor{blue!12} .152 & \cellcolor{blue!34} .429 & \cellcolor{blue!24} .304 & \cellcolor{blue!17} .223 & \cellcolor{blue!25} .321 & 0 & -- & -- \\
MAV & \cellcolor{blue!51} .643 & \cellcolor{blue!44} .562 & \cellcolor{blue!57} .714 & \cellcolor{blue!44} .562 & \cellcolor{blue!49} .616 & \cellcolor{blue!43} .545 & \cellcolor{blue!49} .616 & \cellcolor{blue!26} .330 & \cellcolor{blue!67} .848 & \cellcolor{blue!41} .518 & \cellcolor{blue!32} .411 & \cellcolor{blue!53} .670 & 0 & -- \\
RSD & \cellcolor{blue!29} .366 & \cellcolor{blue!27} .348 & \cellcolor{blue!34} .437 & \cellcolor{blue!27} .348 & \cellcolor{blue!31} .393 & \cellcolor{blue!27} .339 & \cellcolor{blue!24} .304 & \cellcolor{blue!40} .500 & \cellcolor{blue!43} .545 & \cellcolor{blue!27} .348 & \cellcolor{blue!33} .420 & \cellcolor{blue!27} .339 & \cellcolor{blue!49} .616 & 0 \\
Random & \cellcolor{blue!60} .750 & \cellcolor{blue!60} .750 & \cellcolor{blue!56} .705 & \cellcolor{blue!60} .750 & \cellcolor{blue!59} .741 & \cellcolor{blue!61} .768 & \cellcolor{blue!58} .732 & \cellcolor{blue!60} .750 & \cellcolor{blue!56} .705 & \cellcolor{blue!62} .777 & \cellcolor{blue!61} .768 & \cellcolor{blue!60} .750 & \cellcolor{blue!57} .723 & \cellcolor{blue!62} .777 \\
Min & \cellcolor{blue!19} .241 & \cellcolor{blue!17} .223 & \cellcolor{blue!32} .411 & \cellcolor{blue!17} .223 & \cellcolor{blue!17} .223 & \cellcolor{blue!18} .232 & \cellcolor{blue!22} .286 & \cellcolor{blue!27} .348 & \cellcolor{blue!45} .571 & \cellcolor{blue!20} .259 & \cellcolor{blue!21} .268 & \cellcolor{blue!27} .339 & \cellcolor{blue!37} .473 & \cellcolor{blue!32} .402 \\
Max & \cellcolor{blue!75} .946 & \cellcolor{blue!75} .938 & \cellcolor{blue!76} .955 & \cellcolor{blue!75} .938 & \cellcolor{blue!77} .973 & \cellcolor{blue!76} .955 & \cellcolor{blue!76} .955 & \cellcolor{blue!64} .812 & \cellcolor{blue!69} .866 & \cellcolor{blue!75} .938 & \cellcolor{blue!69} .866 & \cellcolor{blue!74} .929 & \cellcolor{blue!57} .714 & \cellcolor{blue!71} .893 \\
NN-all & \cellcolor{blue!32} .409 & \cellcolor{blue!38} .481 & \cellcolor{blue!38} .484 & \cellcolor{blue!38} .481 & \cellcolor{blue!35} .449 & \cellcolor{blue!38} .481 & \cellcolor{blue!35} .445 & \cellcolor{blue!52} .659 & \cellcolor{blue!42} .527 & \cellcolor{blue!39} .499 & \cellcolor{blue!44} .561 & \cellcolor{blue!37} .472 & \cellcolor{blue!59} .748 & \cellcolor{blue!41} .517 \\
\bottomrule
\end{tabular}
\caption{Difference between rules for 5 alternatives with $1 \leq k < 5$ on Real-World preferences.}
\label{tab:rule_distance_heatmap-m=[5]-pref_dist=rwd}
\end{table}

\begin{figure}[htbp!]
\includegraphics[width=0.9\textwidth]{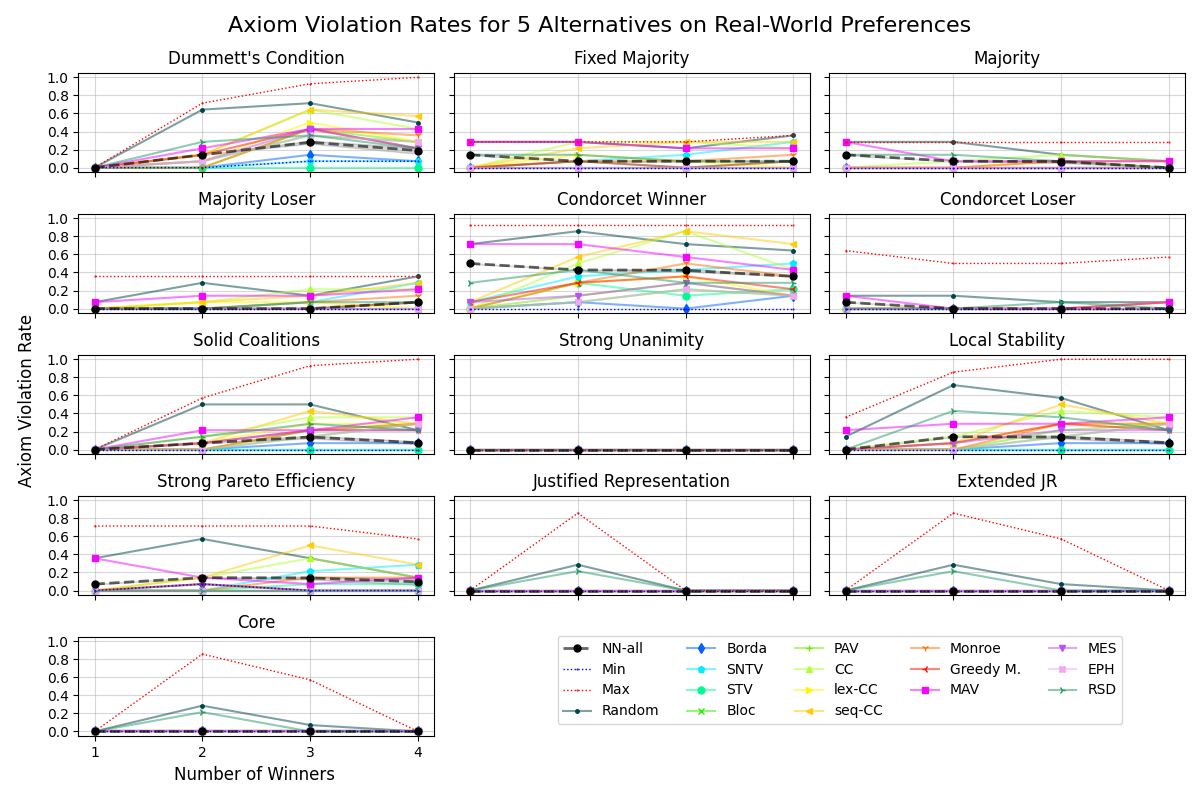}
\caption{Axiom violation rate for each axiom on Real World Data preferences with 5 alternatives.}
\end{figure}

\begin{figure}[htbp!]
\includegraphics[width=0.9\textwidth]{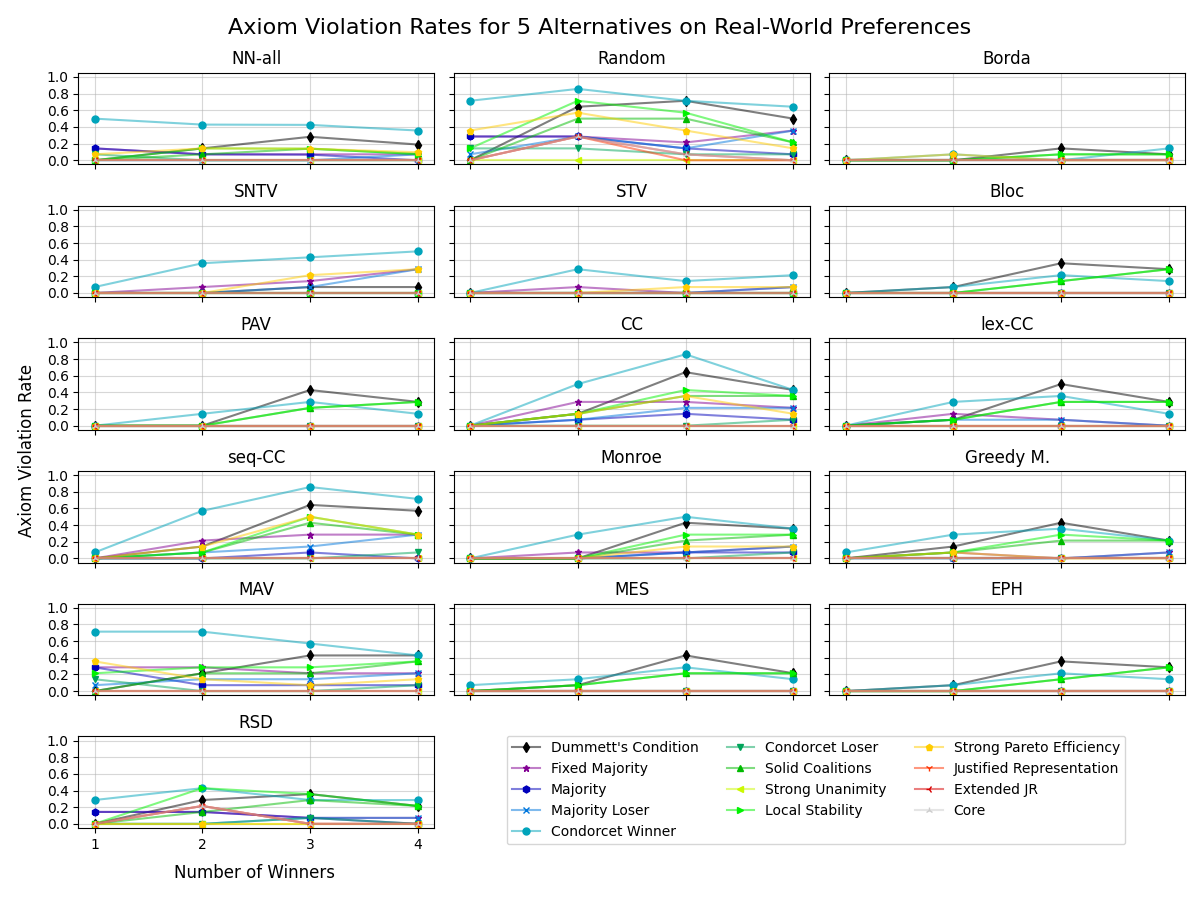}
\caption{Axiom violation rate for each rule on Real World Data preferences with 5 alternatives.}
\end{figure}

\newpage
\clearpage

\subsection{6 Alternatives, Real World Data}
\label{sec:6_alternatives-Real World Data_preferences}
\begin{table}[ht]
\label{tab:summary_table-n_profiles=[25000]-num_voters=[50]-m=[6]-pref_dist=['rwd']-axioms=['all']}
\centering
\fontsize{7pt}{9pt}
\selectfont
\setlength{\tabcolsep}{4.6pt}
\renewcommand{\arraystretch}{1.05}\begin{tabular}{lc|cccccc|ccccccc}
\toprule
Method & \rotatebox{90}{Mean} & \rotatebox{90}{Maj W} & \rotatebox{90}{\underline{Maj L}} & \rotatebox{90}{\underline{Cond W}} & \rotatebox{90}{Cond L} & \rotatebox{90}{\underline{Pareto}} & \rotatebox{90}{F Maj} & \rotatebox{90}{Unanimity} & \rotatebox{90}{\underline{Dummett's}} & \rotatebox{90}{JR} & \rotatebox{90}{EJR} & \rotatebox{90}{\underline{Core}} & \rotatebox{90}{S. Coalitions} & \rotatebox{90}{\underline{Stability}} \\
\midrule
NN-all & .117 & .052 & .033 & .476 & .056 & .101 & .067 & \textbf{0} & .242 & .019 & .030 & .030 & .180 & .241 \\
Min & \textbf{.007} & \textbf{0} & \textbf{0} & \textbf{.022} & \textbf{0} & .022 & \textbf{0} & \textbf{0} & .011 & \textbf{0} & \textbf{0} & \textbf{0} & \textbf{0} & .033 \\
Max & .385 & .222 & .222 & .744 & .511 & .589 & .144 & .011 & .611 & .167 & .211 & .222 & .556 & .789 \\
\midrule
Borda & .023 & .011 & \textbf{0} & .044 & \textbf{0} & .033 & .011 & \cellcolor{green!25}\textbf{0} & .089 & \textbf{0} & \textbf{0} & \textbf{0} & .056 & .056 \\
EPH & .050 & \textbf{0} & \textbf{0} & .222 & \textbf{0} & \cellcolor{green!25}\textbf{0} & \textbf{0} & \textbf{0} & .178 & \textbf{0} & \textbf{0} & \textbf{0} & .122 & .133 \\
SNTV & .030 & \cellcolor{green!25}\textbf{0} & .011 & .311 & \textbf{0} & .044 & .022 & \textbf{0} & \textbf{0} & \textbf{0} & \textbf{0} & \textbf{0} & \cellcolor{green!25}\textbf{0} & \textbf{0} \\
Bloc & .050 & \textbf{0} & \textbf{0} & .200 & \textbf{0} & \cellcolor{green!25}\textbf{0} & \cellcolor{green!25}\textbf{0} & \cellcolor{green!25}\textbf{0} & .178 & \textbf{0} & \textbf{0} & \textbf{0} & .122 & .144 \\
\midrule
STV & .019 & \cellcolor{green!25}\textbf{0} & \textbf{0} & .189 & \textbf{0} & .056 & \textbf{0} & \textbf{0} & \textbf{0} & \textbf{0} & \textbf{0} & \textbf{0} & \cellcolor{green!25}\textbf{0} & \textbf{0} \\
PAV & .056 & \textbf{0} & \textbf{0} & .256 & \textbf{0} & \cellcolor{green!25}\textbf{0} & \textbf{0} & \textbf{0} & .189 & \cellcolor{green!25}\textbf{0} & \cellcolor{green!25}\textbf{0} & \textbf{0} & .144 & .144 \\
MES & .056 & \textbf{0} & \textbf{0} & .256 & \textbf{0} & \textbf{0} & \textbf{0} & \textbf{0} & .178 & \cellcolor{green!25}\textbf{0} & \cellcolor{green!25}\textbf{0} & \textbf{0} & .144 & .144 \\
CC & .159 & .056 & .122 & .511 & .056 & .167 & .100 & .011 & .400 & \cellcolor{green!25}\textbf{0} & .022 & .033 & .256 & .333 \\
seq-CC & .103 & .033 & .022 & .422 & .033 & .067 & .056 & \textbf{0} & .267 & \cellcolor{green!25}\textbf{0} & .011 & .011 & .200 & .211 \\
lex-CC & .072 & \textbf{0} & \textbf{0} & .356 & \textbf{0} & \textbf{0} & .022 & \textbf{0} & .233 & \textbf{0} & \textbf{0} & \textbf{0} & .156 & .167 \\
Monroe & .117 & .033 & .044 & .467 & .033 & .100 & .067 & \cellcolor{green!25}\textbf{0} & .333 & \cellcolor{green!25}\textbf{0} & \textbf{0} & \textbf{0} & .189 & .256 \\
Greedy M. & .064 & .011 & \textbf{0} & .333 & .011 & \textbf{0} & .022 & \cellcolor{green!25}\textbf{0} & .178 & \cellcolor{green!25}\textbf{0} & \textbf{0} & \textbf{0} & .133 & .144 \\
\midrule
MAV & .180 & .078 & .156 & .589 & .089 & .244 & .111 & \textbf{0} & .378 & .022 & .033 & .033 & .222 & .389 \\
RSD & .110 & .078 & .011 & .522 & .022 & \textbf{0} & .078 & \textbf{0} & .244 & .011 & .022 & .022 & .156 & .267 \\
Random & .192 & .122 & .156 & .622 & .100 & .256 & .133 & .011 & .367 & .022 & .067 & .067 & .256 & .322 \\
\bottomrule
\end{tabular}
\caption{Average Axiom Violation Rate for 6 alternatives and $1 \leq k < 6$ winners across Real World Data preferences.}
\end{table}

\begin{table}[ht]
\centering
\fontsize{7pt}{9pt}\selectfont
\setlength{\tabcolsep}{4.6pt}
\renewcommand{\arraystretch}{1.05}
\begin{tabular}{@{}lcccccccccccccc@{}}
\toprule
 & \rotatebox{90}{Borda} & \rotatebox{90}{EPH} & \rotatebox{90}{SNTV} & \rotatebox{90}{Bloc} & \rotatebox{90}{STV} & \rotatebox{90}{PAV} & \rotatebox{90}{MES} & \rotatebox{90}{CC} & \rotatebox{90}{seq-CC} & \rotatebox{90}{lex-CC} & \rotatebox{90}{Monroe} & \rotatebox{90}{Greedy M.} & \rotatebox{90}{MAV} & \rotatebox{90}{RSD} \\
\midrule
Borda & 0 & -- & -- & -- & -- & -- & -- & -- & -- & -- & -- & -- & -- & -- \\
EPH & \cellcolor{blue!24} .306 & 0 & -- & -- & -- & -- & -- & -- & -- & -- & -- & -- & -- & -- \\
SNTV & \cellcolor{blue!33} .415 & \cellcolor{blue!34} .428 & 0 & -- & -- & -- & -- & -- & -- & -- & -- & -- & -- & -- \\
Bloc & \cellcolor{blue!24} .300 & \cellcolor{blue!1} .013 & \cellcolor{blue!34} .431 & 0 & -- & -- & -- & -- & -- & -- & -- & -- & -- & -- \\
STV & \cellcolor{blue!23} .293 & \cellcolor{blue!29} .363 & \cellcolor{blue!14} .185 & \cellcolor{blue!29} .369 & 0 & -- & -- & -- & -- & -- & -- & -- & -- & -- \\
PAV & \cellcolor{blue!26} .326 & \cellcolor{blue!2} .035 & \cellcolor{blue!34} .428 & \cellcolor{blue!3} .044 & \cellcolor{blue!29} .365 & 0 & -- & -- & -- & -- & -- & -- & -- & -- \\
MES & \cellcolor{blue!25} .313 & \cellcolor{blue!13} .165 & \cellcolor{blue!26} .335 & \cellcolor{blue!13} .174 & \cellcolor{blue!28} .350 & \cellcolor{blue!10} .135 & 0 & -- & -- & -- & -- & -- & -- & -- \\
CC & \cellcolor{blue!45} .565 & \cellcolor{blue!37} .467 & \cellcolor{blue!51} .646 & \cellcolor{blue!38} .476 & \cellcolor{blue!46} .578 & \cellcolor{blue!35} .444 & \cellcolor{blue!42} .533 & 0 & -- & -- & -- & -- & -- & -- \\
seq-CC & \cellcolor{blue!38} .476 & \cellcolor{blue!35} .443 & \cellcolor{blue!25} .322 & \cellcolor{blue!35} .448 & \cellcolor{blue!33} .415 & \cellcolor{blue!33} .422 & \cellcolor{blue!25} .317 & \cellcolor{blue!50} .628 & 0 & -- & -- & -- & -- & -- \\
lex-CC & \cellcolor{blue!31} .389 & \cellcolor{blue!8} .107 & \cellcolor{blue!37} .474 & \cellcolor{blue!9} .117 & \cellcolor{blue!32} .406 & \cellcolor{blue!5} .072 & \cellcolor{blue!14} .181 & \cellcolor{blue!33} .424 & \cellcolor{blue!34} .426 & 0 & -- & -- & -- & -- \\
Monroe & \cellcolor{blue!40} .502 & \cellcolor{blue!32} .400 & \cellcolor{blue!46} .583 & \cellcolor{blue!32} .406 & \cellcolor{blue!41} .520 & \cellcolor{blue!30} .381 & \cellcolor{blue!37} .470 & \cellcolor{blue!6} .080 & \cellcolor{blue!47} .596 & \cellcolor{blue!30} .378 & 0 & -- & -- & -- \\
Greedy M. & \cellcolor{blue!30} .387 & \cellcolor{blue!27} .344 & \cellcolor{blue!25} .319 & \cellcolor{blue!28} .354 & \cellcolor{blue!28} .352 & \cellcolor{blue!25} .322 & \cellcolor{blue!17} .217 & \cellcolor{blue!44} .556 & \cellcolor{blue!25} .320 & \cellcolor{blue!26} .331 & \cellcolor{blue!40} .507 & 0 & -- & -- \\
MAV & \cellcolor{blue!54} .687 & \cellcolor{blue!55} .689 & \cellcolor{blue!60} .757 & \cellcolor{blue!55} .693 & \cellcolor{blue!57} .717 & \cellcolor{blue!53} .672 & \cellcolor{blue!54} .683 & \cellcolor{blue!26} .337 & \cellcolor{blue!63} .791 & \cellcolor{blue!52} .654 & \cellcolor{blue!28} .350 & \cellcolor{blue!57} .719 & 0 & -- \\
RSD & \cellcolor{blue!48} .602 & \cellcolor{blue!46} .580 & \cellcolor{blue!48} .607 & \cellcolor{blue!46} .580 & \cellcolor{blue!47} .589 & \cellcolor{blue!45} .570 & \cellcolor{blue!48} .602 & \cellcolor{blue!52} .661 & \cellcolor{blue!54} .676 & \cellcolor{blue!45} .563 & \cellcolor{blue!48} .611 & \cellcolor{blue!47} .591 & \cellcolor{blue!55} .689 & 0 \\
Random & \cellcolor{blue!55} .696 & \cellcolor{blue!56} .704 & \cellcolor{blue!59} .739 & \cellcolor{blue!55} .694 & \cellcolor{blue!57} .722 & \cellcolor{blue!56} .704 & \cellcolor{blue!57} .715 & \cellcolor{blue!57} .713 & \cellcolor{blue!54} .676 & \cellcolor{blue!56} .704 & \cellcolor{blue!55} .696 & \cellcolor{blue!56} .707 & \cellcolor{blue!57} .722 & \cellcolor{blue!54} .687 \\
Min & \cellcolor{blue!17} .220 & \cellcolor{blue!27} .348 & \cellcolor{blue!30} .380 & \cellcolor{blue!27} .346 & \cellcolor{blue!20} .257 & \cellcolor{blue!28} .361 & \cellcolor{blue!30} .385 & \cellcolor{blue!43} .543 & \cellcolor{blue!41} .517 & \cellcolor{blue!32} .404 & \cellcolor{blue!38} .476 & \cellcolor{blue!34} .435 & \cellcolor{blue!47} .596 & \cellcolor{blue!47} .594 \\
Max & \cellcolor{blue!71} .891 & \cellcolor{blue!72} .900 & \cellcolor{blue!73} .920 & \cellcolor{blue!72} .900 & \cellcolor{blue!71} .894 & \cellcolor{blue!71} .889 & \cellcolor{blue!71} .896 & \cellcolor{blue!56} .706 & \cellcolor{blue!70} .876 & \cellcolor{blue!70} .881 & \cellcolor{blue!59} .746 & \cellcolor{blue!71} .898 & \cellcolor{blue!45} .567 & \cellcolor{blue!65} .819 \\
NN-all & \cellcolor{blue!39} .489 & \cellcolor{blue!46} .583 & \cellcolor{blue!38} .484 & \cellcolor{blue!46} .579 & \cellcolor{blue!41} .523 & \cellcolor{blue!46} .580 & \cellcolor{blue!42} .534 & \cellcolor{blue!58} .725 & \cellcolor{blue!34} .428 & \cellcolor{blue!47} .593 & \cellcolor{blue!56} .702 & \cellcolor{blue!42} .535 & \cellcolor{blue!64} .804 & \cellcolor{blue!49} .618 \\
\bottomrule
\end{tabular}
\caption{Difference between rules for 6 alternatives with $1 \leq k < 6$ on Real-World preferences.}
\label{tab:rule_distance_heatmap-m=[6]-pref_dist=rwd}
\end{table}

\begin{figure}[htbp!]
\includegraphics[width=0.9\textwidth]{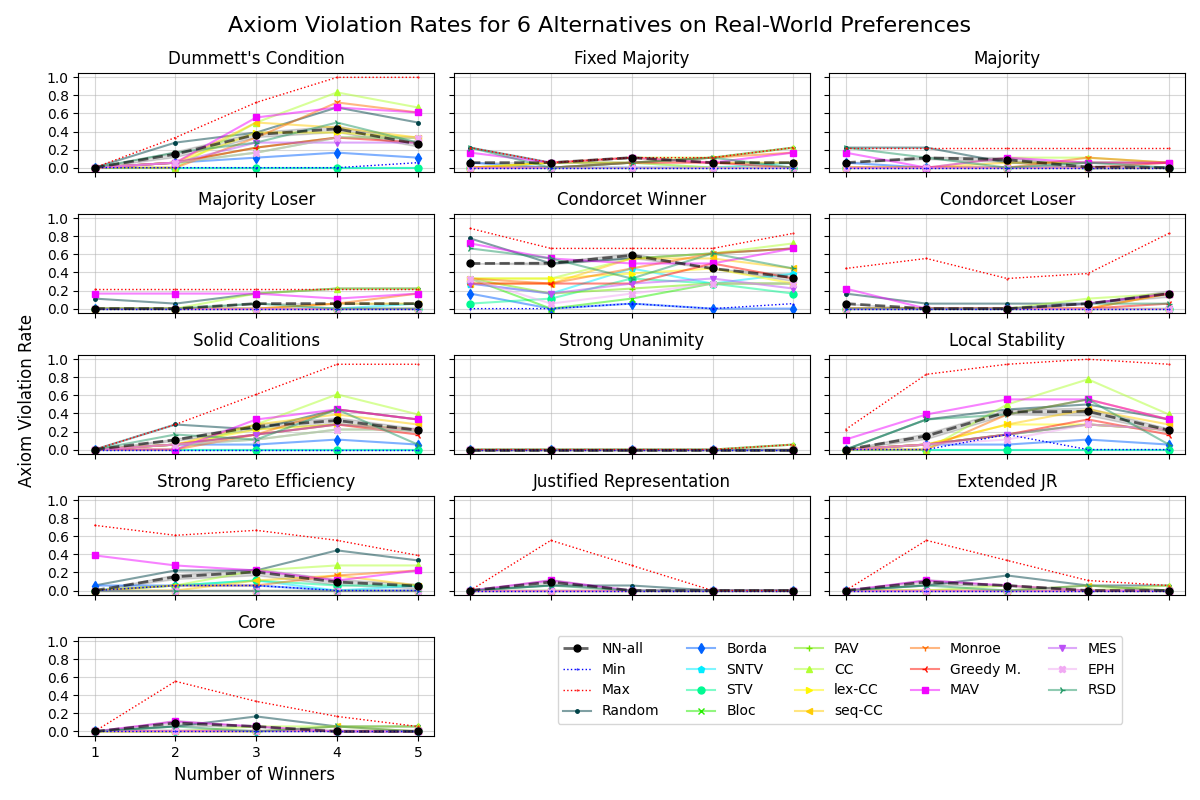}
\caption{Axiom violation rate for each axiom on Real World Data preferences with 6 alternatives.}
\end{figure}

\begin{figure}[htbp!]
\includegraphics[width=0.9\textwidth]{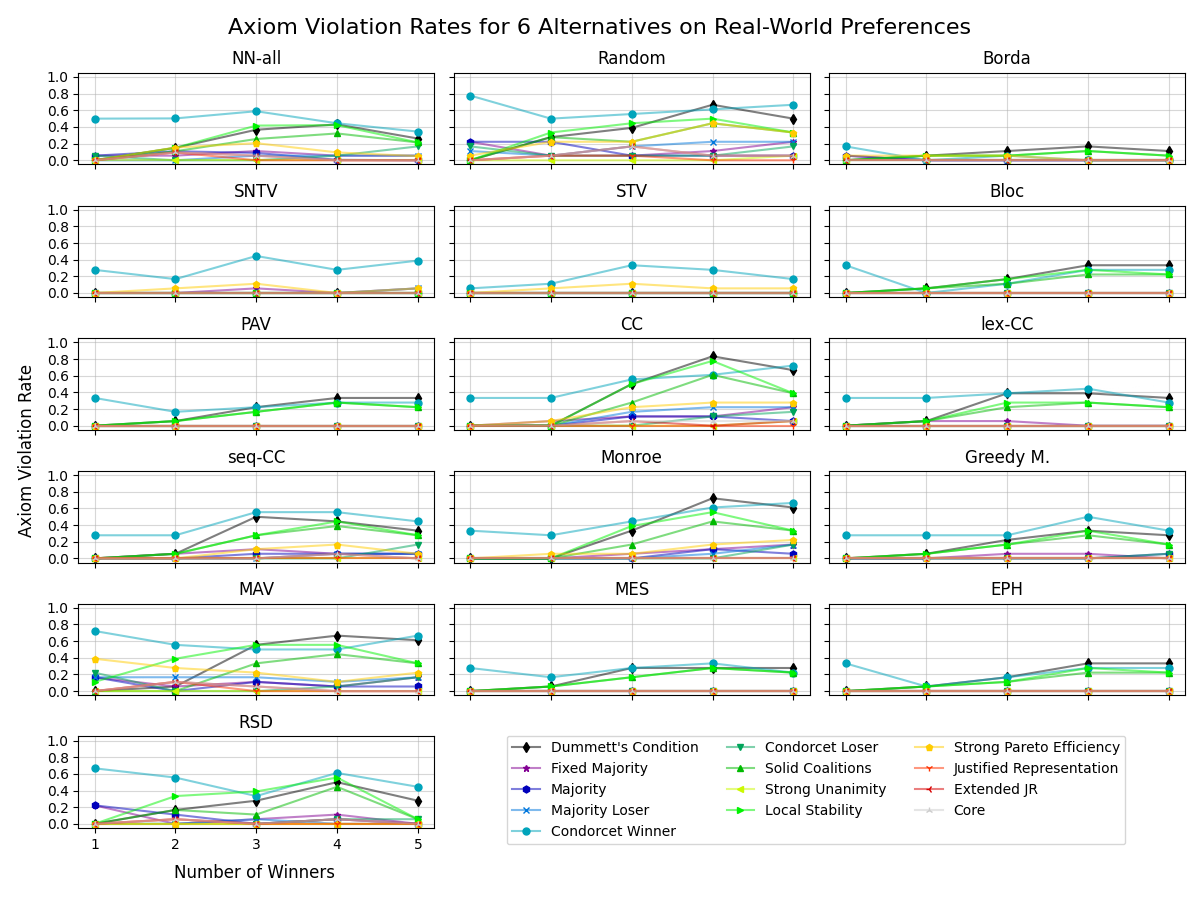}
\caption{Axiom violation rate for each rule on Real World Data preferences with 6 alternatives.}
\end{figure}

\newpage
\clearpage

\subsection{7 Alternatives, Real World Data}
\label{sec:7_alternatives-Real World Data_preferences}
\begin{table}[ht]
\label{tab:summary_table-n_profiles=[25000]-num_voters=[50]-m=[7]-pref_dist=['rwd']-axioms=['all']}
\centering
\fontsize{7pt}{9pt}
\selectfont
\setlength{\tabcolsep}{4.6pt}
\renewcommand{\arraystretch}{1.05}\begin{tabular}{lc|cccccc|ccccccc}
\toprule
Method & \rotatebox{90}{Mean} & \rotatebox{90}{Maj W} & \rotatebox{90}{\underline{Maj L}} & \rotatebox{90}{\underline{Cond W}} & \rotatebox{90}{Cond L} & \rotatebox{90}{\underline{Pareto}} & \rotatebox{90}{F Maj} & \rotatebox{90}{Unanimity} & \rotatebox{90}{\underline{Dummett's}} & \rotatebox{90}{JR} & \rotatebox{90}{EJR} & \rotatebox{90}{\underline{Core}} & \rotatebox{90}{S. Coalitions} & \rotatebox{90}{\underline{Stability}} \\
\midrule
NN-all & .102 & .016 & .039 & .457 & .015 & .108 & .041 & \textbf{0} & .200 & .035 & .042 & .046 & .110 & .223 \\
Min & \textbf{.010} & \textbf{0} & \textbf{0} & \textbf{.068} & \textbf{0} & .008 & \textbf{0} & \textbf{0} & .038 & \textbf{0} & \textbf{0} & \textbf{0} & \textbf{0} & .015 \\
Max & .534 & .409 & .318 & .909 & .712 & .750 & .235 & .061 & .750 & .288 & .432 & .462 & .697 & .917 \\
\midrule
Borda & .030 & .008 & \textbf{0} & .136 & \textbf{0} & \textbf{0} & .015 & \cellcolor{green!25}\textbf{0} & .121 & \textbf{0} & \textbf{0} & \textbf{0} & .045 & .068 \\
EPH & .033 & \textbf{0} & \textbf{0} & .152 & \textbf{0} & \cellcolor{green!25}\textbf{0} & \textbf{0} & \textbf{0} & .136 & \textbf{0} & \textbf{0} & \textbf{0} & .061 & .083 \\
SNTV & .086 & \cellcolor{green!25}\textbf{0} & .061 & .598 & \textbf{0} & .167 & .129 & .023 & .076 & \textbf{0} & .030 & .030 & \cellcolor{green!25}\textbf{0} & \textbf{0} \\
Bloc & .033 & \textbf{0} & \textbf{0} & .144 & \textbf{0} & \cellcolor{green!25}\textbf{0} & \cellcolor{green!25}\textbf{0} & \cellcolor{green!25}\textbf{0} & .129 & \textbf{0} & \textbf{0} & \textbf{0} & .061 & .091 \\
\midrule
STV & .027 & \cellcolor{green!25}\textbf{0} & .015 & .258 & \textbf{0} & .023 & .030 & \textbf{0} & \textbf{.023} & \textbf{0} & \textbf{0} & \textbf{0} & \cellcolor{green!25}\textbf{0} & \textbf{0} \\
PAV & .034 & \textbf{0} & \textbf{0} & .152 & \textbf{0} & \cellcolor{green!25}\textbf{0} & \textbf{0} & \textbf{0} & .144 & \cellcolor{green!25}\textbf{0} & \cellcolor{green!25}\textbf{0} & \textbf{0} & .061 & .083 \\
MES & .046 & \textbf{0} & \textbf{0} & .258 & \textbf{0} & \textbf{0} & \textbf{0} & \textbf{0} & .159 & \cellcolor{green!25}\textbf{0} & \cellcolor{green!25}\textbf{0} & \textbf{0} & .076 & .106 \\
CC & .177 & .053 & .174 & .674 & .008 & .318 & .159 & .038 & .348 & \cellcolor{green!25}\textbf{0} & .091 & .098 & .159 & .182 \\
seq-CC & .184 & .038 & .068 & .712 & .015 & .242 & .152 & .030 & .417 & \cellcolor{green!25}\textbf{0} & .068 & .076 & .258 & .318 \\
lex-CC & .054 & \textbf{0} & \textbf{0} & .326 & \textbf{0} & \textbf{0} & .023 & \textbf{0} & .182 & \textbf{0} & \textbf{0} & \textbf{0} & .076 & .098 \\
Monroe & .097 & .008 & .068 & .545 & \textbf{0} & .152 & .076 & \cellcolor{green!25}\textbf{0} & .235 & \cellcolor{green!25}\textbf{0} & \textbf{0} & \textbf{0} & .076 & .106 \\
Greedy M. & .088 & .008 & .015 & .477 & .008 & .015 & .061 & \cellcolor{green!25}\textbf{0} & .227 & \cellcolor{green!25}\textbf{0} & \textbf{0} & \textbf{0} & .144 & .189 \\
\midrule
MAV & .149 & .068 & .106 & .712 & .038 & .235 & .167 & \textbf{0} & .250 & \textbf{0} & .023 & .023 & .114 & .205 \\
RSD & .091 & .030 & .030 & .523 & \textbf{0} & \textbf{0} & .015 & \textbf{0} & .220 & .008 & .015 & .015 & .144 & .189 \\
Random & .303 & .197 & .152 & .841 & .053 & .523 & .227 & .061 & .500 & .106 & .174 & .220 & .371 & .515 \\
\bottomrule
\end{tabular}
\caption{Average Axiom Violation Rate for 7 alternatives and $1 \leq k < 7$ winners across Real World Data preferences.}
\end{table}

\begin{table}[ht]
\centering
\fontsize{7pt}{9pt}\selectfont
\setlength{\tabcolsep}{4.6pt}
\renewcommand{\arraystretch}{1.05}
\begin{tabular}{@{}lcccccccccccccc@{}}
\toprule
 & \rotatebox{90}{Borda} & \rotatebox{90}{EPH} & \rotatebox{90}{SNTV} & \rotatebox{90}{Bloc} & \rotatebox{90}{STV} & \rotatebox{90}{PAV} & \rotatebox{90}{MES} & \rotatebox{90}{CC} & \rotatebox{90}{seq-CC} & \rotatebox{90}{lex-CC} & \rotatebox{90}{Monroe} & \rotatebox{90}{Greedy M.} & \rotatebox{90}{MAV} & \rotatebox{90}{RSD} \\
\midrule
Borda & 0 & -- & -- & -- & -- & -- & -- & -- & -- & -- & -- & -- & -- & -- \\
EPH & \cellcolor{blue!12} .150 & 0 & -- & -- & -- & -- & -- & -- & -- & -- & -- & -- & -- & -- \\
SNTV & \cellcolor{blue!33} .422 & \cellcolor{blue!32} .409 & 0 & -- & -- & -- & -- & -- & -- & -- & -- & -- & -- & -- \\
Bloc & \cellcolor{blue!12} .154 & \cellcolor{blue!0} .008 & \cellcolor{blue!33} .413 & 0 & -- & -- & -- & -- & -- & -- & -- & -- & -- & -- \\
STV & \cellcolor{blue!18} .235 & \cellcolor{blue!20} .254 & \cellcolor{blue!26} .333 & \cellcolor{blue!20} .254 & 0 & -- & -- & -- & -- & -- & -- & -- & -- & -- \\
PAV & \cellcolor{blue!12} .152 & \cellcolor{blue!2} .029 & \cellcolor{blue!33} .419 & \cellcolor{blue!2} .033 & \cellcolor{blue!20} .256 & 0 & -- & -- & -- & -- & -- & -- & -- & -- \\
MES & \cellcolor{blue!16} .203 & \cellcolor{blue!10} .129 & \cellcolor{blue!29} .365 & \cellcolor{blue!10} .133 & \cellcolor{blue!25} .321 & \cellcolor{blue!9} .121 & 0 & -- & -- & -- & -- & -- & -- & -- \\
CC & \cellcolor{blue!42} .527 & \cellcolor{blue!36} .455 & \cellcolor{blue!46} .581 & \cellcolor{blue!36} .458 & \cellcolor{blue!39} .496 & \cellcolor{blue!35} .449 & \cellcolor{blue!41} .520 & 0 & -- & -- & -- & -- & -- & -- \\
seq-CC & \cellcolor{blue!41} .520 & \cellcolor{blue!39} .491 & \cellcolor{blue!29} .367 & \cellcolor{blue!39} .495 & \cellcolor{blue!43} .545 & \cellcolor{blue!38} .475 & \cellcolor{blue!32} .409 & \cellcolor{blue!57} .722 & 0 & -- & -- & -- & -- & -- \\
lex-CC & \cellcolor{blue!17} .220 & \cellcolor{blue!9} .119 & \cellcolor{blue!34} .426 & \cellcolor{blue!9} .122 & \cellcolor{blue!23} .295 & \cellcolor{blue!7} .092 & \cellcolor{blue!11} .145 & \cellcolor{blue!34} .431 & \cellcolor{blue!37} .472 & 0 & -- & -- & -- & -- \\
Monroe & \cellcolor{blue!34} .433 & \cellcolor{blue!28} .352 & \cellcolor{blue!41} .514 & \cellcolor{blue!28} .356 & \cellcolor{blue!31} .396 & \cellcolor{blue!27} .342 & \cellcolor{blue!33} .415 & \cellcolor{blue!10} .133 & \cellcolor{blue!50} .628 & \cellcolor{blue!26} .328 & 0 & -- & -- & -- \\
Greedy M. & \cellcolor{blue!27} .345 & \cellcolor{blue!23} .298 & \cellcolor{blue!30} .376 & \cellcolor{blue!24} .306 & \cellcolor{blue!29} .369 & \cellcolor{blue!23} .293 & \cellcolor{blue!16} .212 & \cellcolor{blue!47} .597 & \cellcolor{blue!29} .374 & \cellcolor{blue!24} .306 & \cellcolor{blue!39} .489 & 0 & -- & -- \\
MAV & \cellcolor{blue!43} .539 & \cellcolor{blue!42} .525 & \cellcolor{blue!52} .652 & \cellcolor{blue!42} .525 & \cellcolor{blue!45} .566 & \cellcolor{blue!41} .520 & \cellcolor{blue!44} .557 & \cellcolor{blue!24} .312 & \cellcolor{blue!61} .770 & \cellcolor{blue!39} .497 & \cellcolor{blue!26} .332 & \cellcolor{blue!51} .646 & 0 & -- \\
RSD & \cellcolor{blue!32} .400 & \cellcolor{blue!29} .367 & \cellcolor{blue!41} .515 & \cellcolor{blue!29} .367 & \cellcolor{blue!31} .396 & \cellcolor{blue!30} .376 & \cellcolor{blue!31} .393 & \cellcolor{blue!44} .556 & \cellcolor{blue!45} .571 & \cellcolor{blue!31} .390 & \cellcolor{blue!37} .465 & \cellcolor{blue!36} .461 & \cellcolor{blue!44} .556 & 0 \\
Random & \cellcolor{blue!60} .758 & \cellcolor{blue!58} .732 & \cellcolor{blue!57} .717 & \cellcolor{blue!58} .732 & \cellcolor{blue!57} .721 & \cellcolor{blue!59} .739 & \cellcolor{blue!59} .739 & \cellcolor{blue!61} .773 & \cellcolor{blue!58} .725 & \cellcolor{blue!59} .739 & \cellcolor{blue!61} .768 & \cellcolor{blue!60} .753 & \cellcolor{blue!59} .739 & \cellcolor{blue!58} .725 \\
Min & \cellcolor{blue!14} .181 & \cellcolor{blue!15} .199 & \cellcolor{blue!34} .427 & \cellcolor{blue!15} .196 & \cellcolor{blue!13} .172 & \cellcolor{blue!15} .198 & \cellcolor{blue!21} .265 & \cellcolor{blue!37} .473 & \cellcolor{blue!46} .586 & \cellcolor{blue!20} .253 & \cellcolor{blue!30} .383 & \cellcolor{blue!32} .400 & \cellcolor{blue!40} .508 & \cellcolor{blue!32} .400 \\
Max & \cellcolor{blue!77} .970 & \cellcolor{blue!78} .975 & \cellcolor{blue!73} .914 & \cellcolor{blue!77} .971 & \cellcolor{blue!74} .934 & \cellcolor{blue!77} .972 & \cellcolor{blue!78} .977 & \cellcolor{blue!62} .778 & \cellcolor{blue!67} .848 & \cellcolor{blue!75} .944 & \cellcolor{blue!68} .854 & \cellcolor{blue!74} .937 & \cellcolor{blue!63} .789 & \cellcolor{blue!72} .900 \\
NN-all & \cellcolor{blue!34} .427 & \cellcolor{blue!35} .440 & \cellcolor{blue!40} .507 & \cellcolor{blue!34} .436 & \cellcolor{blue!35} .444 & \cellcolor{blue!35} .440 & \cellcolor{blue!36} .453 & \cellcolor{blue!55} .693 & \cellcolor{blue!39} .492 & \cellcolor{blue!35} .447 & \cellcolor{blue!49} .617 & \cellcolor{blue!37} .465 & \cellcolor{blue!52} .656 & \cellcolor{blue!40} .509 \\
\bottomrule
\end{tabular}
\caption{Difference between rules for 7 alternatives with $1 \leq k < 7$ on Real-World preferences.}
\label{tab:rule_distance_heatmap-m=[7]-pref_dist=rwd}
\end{table}

\begin{figure}[htbp!]
\includegraphics[width=0.9\textwidth]{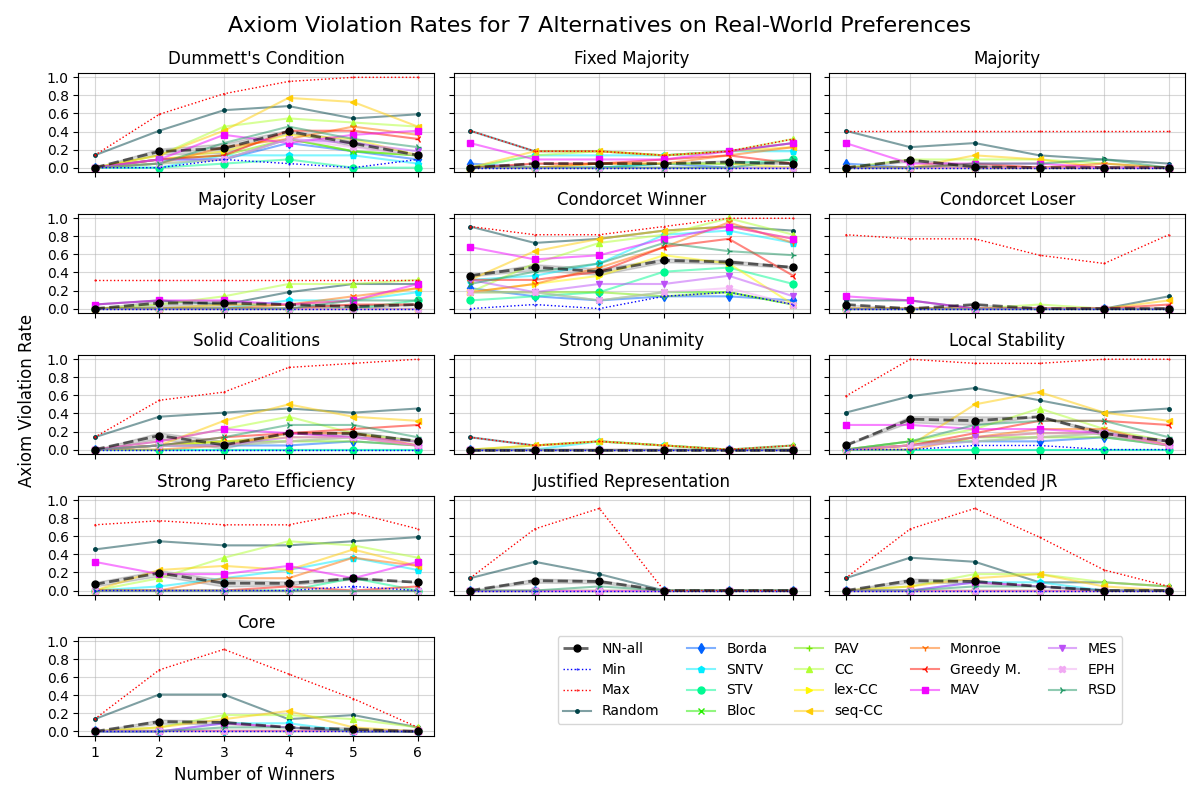}
\caption{Axiom violation rate for each axiom on Real World Data preferences with 7 alternatives.}
\end{figure}

\begin{figure}[htbp!]
\includegraphics[width=0.9\textwidth]{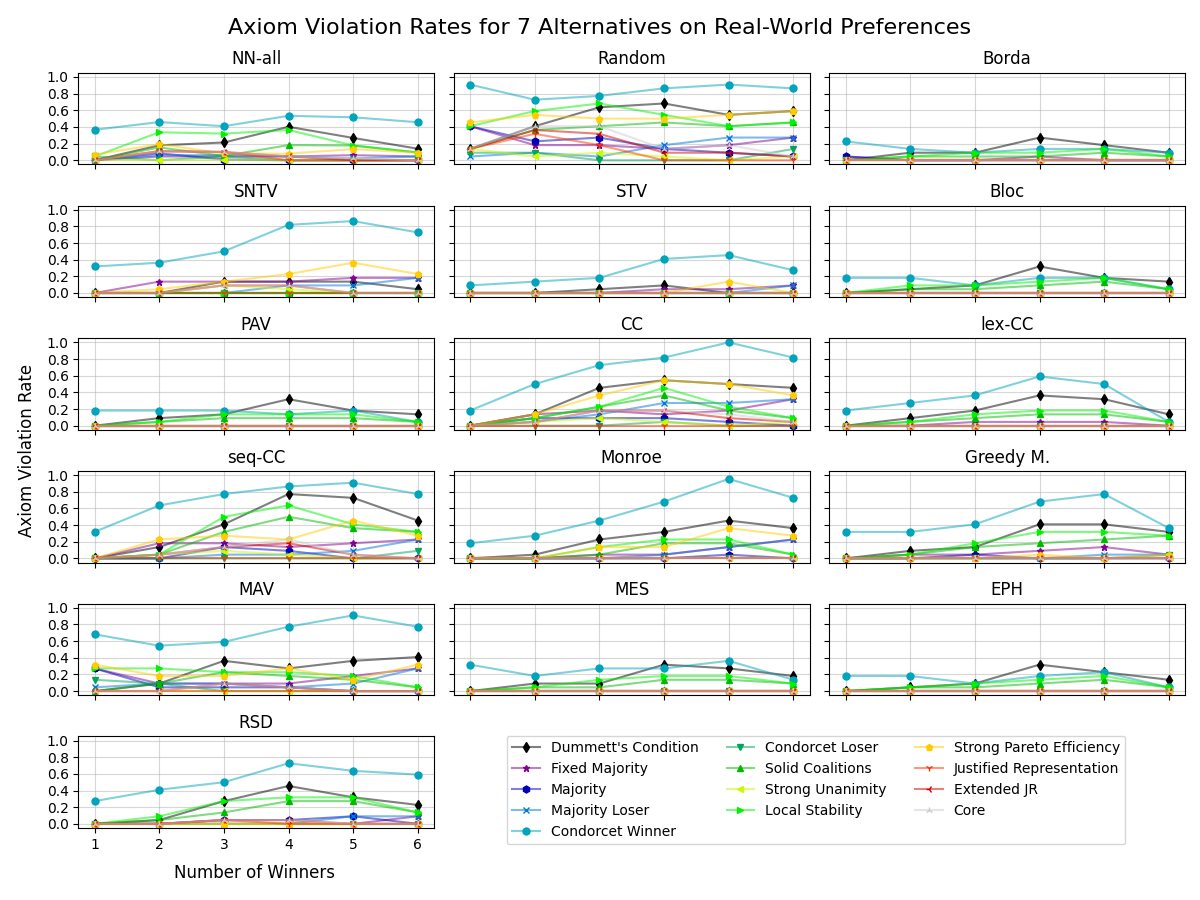}
\caption{Axiom violation rate for each rule on Real World Data preferences with 7 alternatives.}
\end{figure}

\newpage
\clearpage

\section{Additional Experimental Results}
\label{app:additional_experiment_results}

This section includes some additional results for our experiments: the figures matching the body of our paper corresponding to $m = 5$ and $m=6$ alternatives, as well as exact numerical results for the optimal positional scoring rule experiment.

\subsection{Optimized Positional Scoring Rules}
\label{app:optimized_psr}

\autoref{tab:appendix-annealed-psr-complete} shows results of annealing for 1000 steps using 2000 profiles sampled from the Mixed distribution, for each axiom set. Each vector is then evaluated on a 25000 profiles also sampled from the Mixed distribution. The score vector resulting from annealing is shown in the final column.

\begin{table}[h]
\centering
\begin{tabular}{@{}cccc@{}}
\midrule
\multicolumn{4}{c}{\textbf{All Axioms}}                              \\ \midrule
$k$ & \rl{Borda} & \rl{Opt} & Annealed Vector                        \\ \midrule
1   & 0.010      & 0.008    & $(1, 0.69, 0.51, 0.30, 0.08, 0.05, 0)$ \\
2   & 0.011      & 0.011    & $(1, 0.82, 0.50, 0.45, 0.10, 0.01, 0)$ \\
3   & 0.022      & 0.018    & $(1, 0.77, 0.71, 0.24, 0.16, 0.08, 0)$ \\
4   & 0.030      & 0.025    & $(1, 0.74, 0.51, 0.46, 0.12, 0.09, 0)$ \\
5   & 0.034      & 0.028    & $(1, 0.50, 0.44, 0.32, 0.25, 0.05, 0)$ \\
6   & 0.025      & 0.022    & $(1, 0.54, 0.42, 0.42, 0.37, 0.30, 0)$ \\ \midrule
\multicolumn{4}{c}{\textbf{Reduced Axioms}}                          \\ \midrule
$k$ & \rl{Borda} & \rl{Opt} & Annealed Vector                        \\
1   & 0.019      & 0.016    & $(1, 0.70, 0.42, 0.36, 0.19, 0.05, 0)$ \\
2   & 0.021      & 0.018    & $(1, 0.87, 0.54, 0.41, 0.26, 0.01, 0)$ \\
3   & 0.036      & 0.032    & $(1, 0.85, 0.70, 0.38, 0.13, 0.02, 0)$ \\
4   & 0.044      & 0.043    & $(1, 0.63, 0.60, 0.51, 0.22, 0.05, 0)$ \\
5   & 0.042      & 0.042    & $(1, 0.55, 0.47, 0.41, 0.38, 0.08, 0)$ \\
6   & 0.030      & 0.027    & $(1, 0.71, 0.50, 0.48, 0.43, 0.31, 0)$ \\ \bottomrule
\end{tabular}
\caption{Score vector resulting from annealing and the axiom violation rate across a test set of 25000 profiles for $m=7$ alternatives on both axiom sets by \rl{Borda} and the positional scoring rule found via simulated annealing, \rl{Opt}.}
\label{tab:appendix-annealed-psr-complete}
\end{table}

\subsection{Results for 5 Alternatives}
\label{app:5_agg}
\begin{figure*}[ht]
    \includegraphics[width=\textwidth]{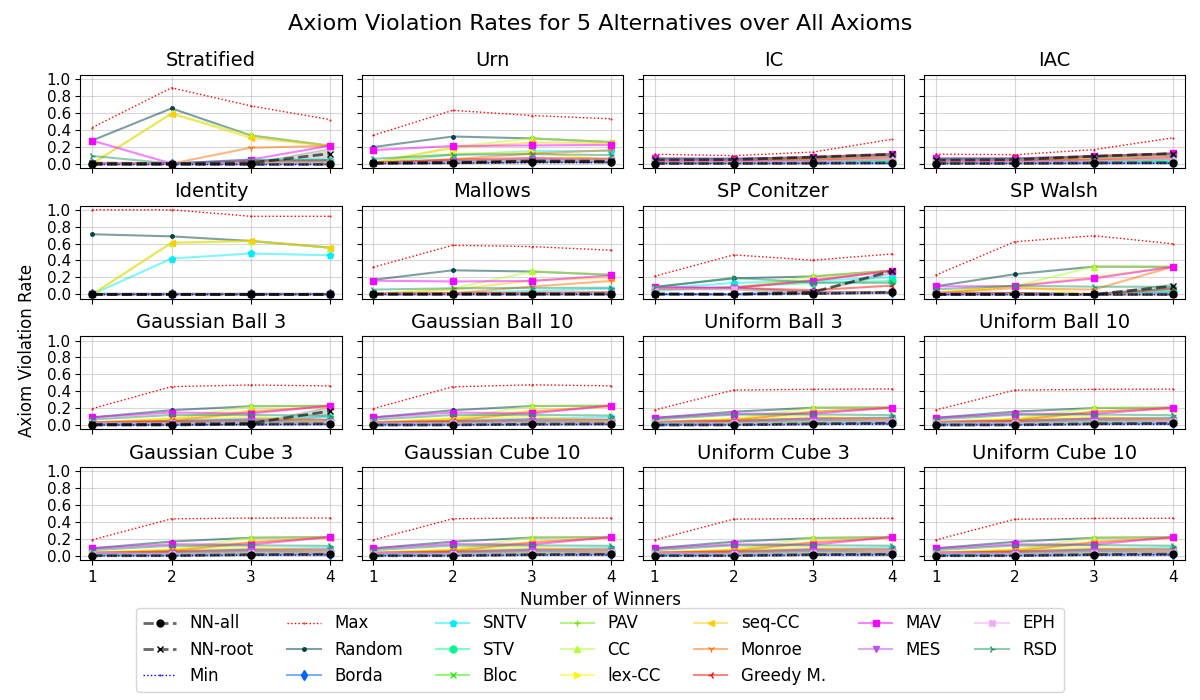}
    \caption{Axiom violation rates for each rule under each individual preference distribution for $m=5$. In all cases, $\mathcal{F}^\text{NN-all}$ has AVR lower than, or similar to, other rules.
    }
    \label{fig:violations_by_distribution-m=5}
\end{figure*}

\begin{table}[ht]
\label{tab:summary_table-n_profiles=[25000]-num_voters=[50]-m=[5]-pref_dist=['all']-axioms=['both']}
\centering
\fontsize{7pt}{9pt}
\selectfont
\setlength{\tabcolsep}{4.6pt}
\renewcommand{\arraystretch}{1.05}\begin{tabular}{lc|cccccc|ccccccc}
\toprule
Method & \rotatebox{90}{Mean} & \rotatebox{90}{Maj W} & \rotatebox{90}{\underline{Maj L}} & \rotatebox{90}{\underline{Cond W}} & \rotatebox{90}{Cond L} & \rotatebox{90}{\underline{Pareto}} & \rotatebox{90}{F Maj} & \rotatebox{90}{Unanimity} & \rotatebox{90}{\underline{Dummett's}} & \rotatebox{90}{JR} & \rotatebox{90}{EJR} & \rotatebox{90}{\underline{Core}} & \rotatebox{90}{S. Coalitions} & \rotatebox{90}{\underline{Stability}} \\
\midrule
NN-all & .006 & \textit{.000} & \textit{.000} & \textbf{.001} & \textit{.000} & .002 & \textit{.000} & \textbf{0} & .027 & \textit{.000} & \textit{.000} & \textit{.000} & .017 & .025 \\
NN-root & .022 & .002 & .020 & .131 & .013 & .017 & .023 & \textbf{0} & .030 & \textit{.000} & \textit{.000} & \textit{.000} & .021 & .028 \\
Min & \textbf{.003} & \textbf{0} & \textit{.000} & .013 & \textit{.000} & \textit{.000} & \textit{.000} & \textbf{0} & .009 & \textbf{0} & \textbf{0} & \textbf{0} & .004 & .012 \\
Max & .430 & .218 & .425 & .959 & .753 & .458 & .268 & .082 & .478 & .196 & .324 & .324 & .443 & .665 \\
\midrule
Borda & .012 & .002 & .007 & .096 & \textbf{0} & .002 & .016 & \cellcolor{green!25}\textbf{0} & .015 & \textit{.000} & \textit{.000} & \textit{.000} & .009 & .011 \\
EPH & .025 & \textit{.000} & .003 & .205 & .005 & \cellcolor{green!25}\textbf{0} & .001 & \textbf{0} & .044 & \textit{.000} & \textit{.000} & \textit{.000} & .030 & .038 \\
SNTV & .078 & \cellcolor{green!25}\textbf{0} & .109 & .484 & .016 & .132 & .111 & .038 & .044 & \textit{.000} & .041 & .041 & \cellcolor{green!25}\textbf{0} & .003 \\
Bloc & .024 & \textit{.000} & .003 & .197 & .005 & \cellcolor{green!25}\textbf{0} & \cellcolor{green!25}\textbf{0} & \cellcolor{green!25}\textbf{0} & .043 & \textit{.000} & \textit{.000} & \textit{.000} & .030 & .040 \\
\midrule
STV & .034 & \cellcolor{green!25}\textbf{0} & .042 & .298 & .005 & .053 & .039 & \textbf{0} & \textbf{0} & \textit{.000} & \textit{.000} & \textit{.000} & \cellcolor{green!25}\textbf{0} & \textit{.000} \\
PAV & .029 & .001 & .003 & .242 & .005 & \cellcolor{green!25}\textbf{0} & .006 & \textbf{0} & .046 & \cellcolor{green!25}\textbf{0} & \cellcolor{green!25}\textbf{0} & \textbf{0} & .033 & .036 \\
MES & .031 & .001 & .004 & .256 & .005 & .001 & .009 & \textbf{0} & .049 & \cellcolor{green!25}\textbf{0} & \cellcolor{green!25}\textbf{0} & \textbf{0} & .035 & .037 \\
CC & .164 & .054 & .163 & .660 & .069 & .217 & .179 & .057 & .230 & \cellcolor{green!25}\textbf{0} & .071 & .071 & .166 & .196 \\
seq-CC & .148 & .046 & .156 & .627 & .044 & .182 & .175 & .057 & .208 & \cellcolor{green!25}\textbf{0} & .059 & .059 & .145 & .167 \\
lex-CC & .047 & .010 & .013 & .362 & .005 & \textbf{0} & .033 & \textbf{0} & .072 & \textbf{0} & \textbf{0} & \textbf{0} & .055 & .057 \\
Monroe & .098 & .016 & .090 & .525 & .048 & .125 & .086 & \cellcolor{green!25}\textbf{0} & .136 & \cellcolor{green!25}\textbf{0} & .001 & .001 & .111 & .130 \\
Greedy M. & .045 & .002 & .026 & .348 & .006 & .006 & .033 & \cellcolor{green!25}\textbf{0} & .063 & \cellcolor{green!25}\textbf{0} & \textbf{0} & \textbf{0} & .046 & .051 \\
\midrule
MAV & .134 & .054 & .135 & .700 & .096 & .128 & .150 & \textbf{0} & .147 & .013 & .015 & .015 & .118 & .176 \\
RSD & .089 & .023 & .085 & .536 & .040 & \textbf{0} & .070 & \textbf{0} & .095 & .023 & .024 & .024 & .074 & .161 \\
Random & .218 & .109 & .215 & .804 & .127 & .285 & .225 & .071 & .250 & .050 & .111 & .112 & .191 & .287 \\
\bottomrule
\end{tabular}
\caption{Average Axiom Violation Rate for 5 alternatives and $1 \leq k < 5$ winners across all preferences.}
\end{table}

\begin{figure}[ht]
    \centering
    \includegraphics[width=\linewidth]{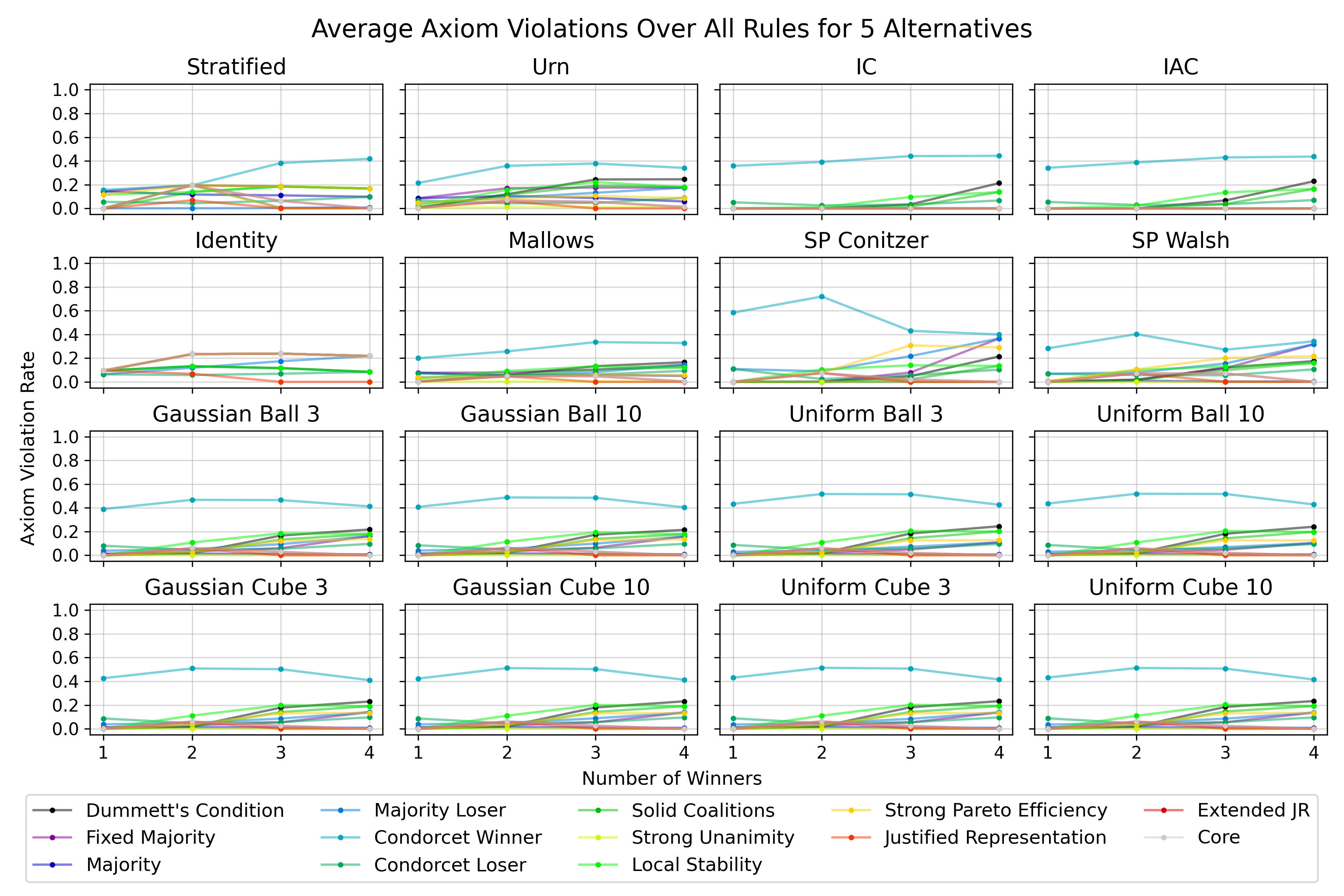}
    \caption{
    Violation rates of each axiom for every preference distribution, averaged over all rules. This shows trends common across all rules in the relationships between axioms and preference distributions.
    }
    \label{fig:violations_by_rule-m=5}
\end{figure}

\begin{table}[ht]
\centering
\fontsize{7pt}{9pt}\selectfont
\setlength{\tabcolsep}{4.6pt}
\renewcommand{\arraystretch}{1.05}
\begin{tabular}{@{}lcccccccccccccc@{}}
\toprule
 & \rotatebox{90}{Borda} & \rotatebox{90}{EPH} & \rotatebox{90}{SNTV} & \rotatebox{90}{Bloc} & \rotatebox{90}{STV} & \rotatebox{90}{PAV} & \rotatebox{90}{MES} & \rotatebox{90}{CC} & \rotatebox{90}{seq-CC} & \rotatebox{90}{lex-CC} & \rotatebox{90}{Monroe} & \rotatebox{90}{Greedy M.} & \rotatebox{90}{MAV} & \rotatebox{90}{RSD} \\
\midrule
Borda & 0 & -- & -- & -- & -- & -- & -- & -- & -- & -- & -- & -- & -- & -- \\
EPH & \cellcolor{blue!18} .227 & 0 & -- & -- & -- & -- & -- & -- & -- & -- & -- & -- & -- & -- \\
SNTV & \cellcolor{blue!31} .394 & \cellcolor{blue!28} .355 & 0 & -- & -- & -- & -- & -- & -- & -- & -- & -- & -- & -- \\
Bloc & \cellcolor{blue!18} .229 & \cellcolor{blue!0} .011 & \cellcolor{blue!28} .353 & 0 & -- & -- & -- & -- & -- & -- & -- & -- & -- & -- \\
STV & \cellcolor{blue!19} .245 & \cellcolor{blue!24} .303 & \cellcolor{blue!20} .261 & \cellcolor{blue!24} .303 & 0 & -- & -- & -- & -- & -- & -- & -- & -- & -- \\
PAV & \cellcolor{blue!18} .235 & \cellcolor{blue!3} .045 & \cellcolor{blue!29} .365 & \cellcolor{blue!4} .053 & \cellcolor{blue!24} .309 & 0 & -- & -- & -- & -- & -- & -- & -- & -- \\
MES & \cellcolor{blue!19} .239 & \cellcolor{blue!7} .096 & \cellcolor{blue!27} .344 & \cellcolor{blue!8} .103 & \cellcolor{blue!25} .315 & \cellcolor{blue!5} .066 & 0 & -- & -- & -- & -- & -- & -- & -- \\
CC & \cellcolor{blue!43} .549 & \cellcolor{blue!34} .429 & \cellcolor{blue!41} .524 & \cellcolor{blue!34} .432 & \cellcolor{blue!41} .521 & \cellcolor{blue!32} .405 & \cellcolor{blue!35} .439 & 0 & -- & -- & -- & -- & -- & -- \\
seq-CC & \cellcolor{blue!40} .507 & \cellcolor{blue!33} .424 & \cellcolor{blue!32} .409 & \cellcolor{blue!34} .429 & \cellcolor{blue!40} .505 & \cellcolor{blue!32} .405 & \cellcolor{blue!29} .367 & \cellcolor{blue!46} .586 & 0 & -- & -- & -- & -- & -- \\
lex-CC & \cellcolor{blue!23} .293 & \cellcolor{blue!9} .119 & \cellcolor{blue!31} .397 & \cellcolor{blue!10} .125 & \cellcolor{blue!28} .350 & \cellcolor{blue!6} .079 & \cellcolor{blue!9} .116 & \cellcolor{blue!29} .367 & \cellcolor{blue!31} .396 & 0 & -- & -- & -- & -- \\
Monroe & \cellcolor{blue!36} .452 & \cellcolor{blue!26} .329 & \cellcolor{blue!36} .453 & \cellcolor{blue!26} .332 & \cellcolor{blue!34} .427 & \cellcolor{blue!24} .310 & \cellcolor{blue!27} .345 & \cellcolor{blue!9} .124 & \cellcolor{blue!43} .541 & \cellcolor{blue!25} .318 & 0 & -- & -- & -- \\
Greedy M. & \cellcolor{blue!24} .301 & \cellcolor{blue!15} .195 & \cellcolor{blue!29} .363 & \cellcolor{blue!16} .201 & \cellcolor{blue!27} .348 & \cellcolor{blue!13} .171 & \cellcolor{blue!11} .140 & \cellcolor{blue!35} .444 & \cellcolor{blue!27} .343 & \cellcolor{blue!15} .194 & \cellcolor{blue!29} .365 & 0 & -- & -- \\
MAV & \cellcolor{blue!47} .594 & \cellcolor{blue!46} .576 & \cellcolor{blue!53} .667 & \cellcolor{blue!46} .577 & \cellcolor{blue!47} .595 & \cellcolor{blue!45} .563 & \cellcolor{blue!46} .582 & \cellcolor{blue!30} .376 & \cellcolor{blue!59} .743 & \cellcolor{blue!42} .528 & \cellcolor{blue!28} .352 & \cellcolor{blue!47} .594 & 0 & -- \\
RSD & \cellcolor{blue!37} .472 & \cellcolor{blue!36} .450 & \cellcolor{blue!44} .553 & \cellcolor{blue!36} .450 & \cellcolor{blue!39} .499 & \cellcolor{blue!36} .453 & \cellcolor{blue!36} .454 & \cellcolor{blue!49} .614 & \cellcolor{blue!47} .593 & \cellcolor{blue!37} .471 & \cellcolor{blue!43} .542 & \cellcolor{blue!37} .467 & \cellcolor{blue!48} .610 & 0 \\
Random & \cellcolor{blue!56} .700 & \cellcolor{blue!56} .700 & \cellcolor{blue!56} .700 & \cellcolor{blue!56} .700 & \cellcolor{blue!56} .700 & \cellcolor{blue!56} .700 & \cellcolor{blue!56} .700 & \cellcolor{blue!56} .700 & \cellcolor{blue!56} .700 & \cellcolor{blue!56} .700 & \cellcolor{blue!56} .700 & \cellcolor{blue!56} .700 & \cellcolor{blue!56} .700 & \cellcolor{blue!56} .700 \\
Min & \cellcolor{blue!10} .125 & \cellcolor{blue!16} .211 & \cellcolor{blue!32} .407 & \cellcolor{blue!16} .207 & \cellcolor{blue!19} .249 & \cellcolor{blue!18} .230 & \cellcolor{blue!19} .246 & \cellcolor{blue!41} .521 & \cellcolor{blue!42} .525 & \cellcolor{blue!23} .289 & \cellcolor{blue!33} .424 & \cellcolor{blue!24} .310 & \cellcolor{blue!44} .562 & \cellcolor{blue!37} .470 \\
Max & \cellcolor{blue!77} .964 & \cellcolor{blue!75} .948 & \cellcolor{blue!71} .897 & \cellcolor{blue!76} .951 & \cellcolor{blue!74} .936 & \cellcolor{blue!74} .935 & \cellcolor{blue!74} .930 & \cellcolor{blue!64} .808 & \cellcolor{blue!62} .785 & \cellcolor{blue!72} .910 & \cellcolor{blue!69} .867 & \cellcolor{blue!72} .911 & \cellcolor{blue!65} .821 & \cellcolor{blue!68} .852 \\
NN-all & \cellcolor{blue!10} .125 & \cellcolor{blue!16} .201 & \cellcolor{blue!33} .413 & \cellcolor{blue!15} .197 & \cellcolor{blue!20} .256 & \cellcolor{blue!17} .221 & \cellcolor{blue!19} .239 & \cellcolor{blue!41} .522 & \cellcolor{blue!41} .524 & \cellcolor{blue!22} .281 & \cellcolor{blue!34} .425 & \cellcolor{blue!24} .306 & \cellcolor{blue!45} .563 & \cellcolor{blue!37} .468 \\
NN-root & \cellcolor{blue!19} .248 & \cellcolor{blue!22} .286 & \cellcolor{blue!35} .443 & \cellcolor{blue!22} .285 & \cellcolor{blue!25} .314 & \cellcolor{blue!23} .295 & \cellcolor{blue!24} .309 & \cellcolor{blue!38} .480 & \cellcolor{blue!44} .555 & \cellcolor{blue!26} .333 & \cellcolor{blue!27} .348 & \cellcolor{blue!27} .342 & \cellcolor{blue!36} .453 & \cellcolor{blue!36} .456 \\
\bottomrule
\end{tabular}

\caption{Difference between rules for 5 alternatives with $1 \leq k < 5$ averaged over all preference distributions.}
\label{tab:rule_distance_heatmap-m=[5]-pref_dist=all}
\end{table}

\newpage
\clearpage

\subsection{Results for 6 Alternatives}
\label{app:6_agg}
\begin{figure*}[ht]
    \includegraphics[width=\textwidth]{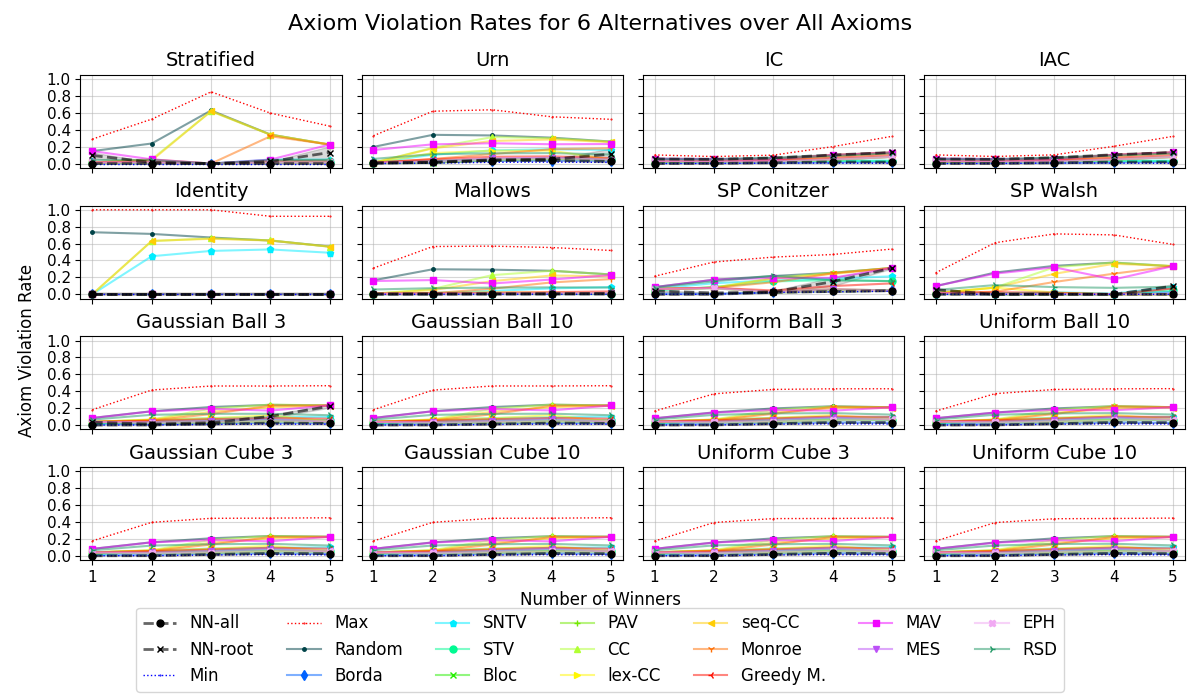}
    \caption{Axiom violation rates for each rule under each individual preference distribution for $m=6$. In all cases, $\mathcal{F}^\text{NN-all}$ has AVR lower than, or similar to, other rules.
    }
    \label{fig:violations_by_distribution-m=6}
\end{figure*}

\begin{table}[ht]
\label{tab:summary_table-n_profiles=[25000]-num_voters=[50]-m=[6]-pref_dist=['all']-axioms=['both']}
\centering
\fontsize{7pt}{9pt}
\selectfont
\setlength{\tabcolsep}{4.6pt}
\renewcommand{\arraystretch}{1.05}\begin{tabular}{lc|cccccc|ccccccc}
\toprule
Method & \rotatebox{90}{Mean} & \rotatebox{90}{Maj W} & \rotatebox{90}{\underline{Maj L}} & \rotatebox{90}{\underline{Cond W}} & \rotatebox{90}{Cond L} & \rotatebox{90}{\underline{Pareto}} & \rotatebox{90}{F Maj} & \rotatebox{90}{Unanimity} & \rotatebox{90}{\underline{Dummett's}} & \rotatebox{90}{JR} & \rotatebox{90}{EJR} & \rotatebox{90}{\underline{Core}} & \rotatebox{90}{S. Coalitions} & \rotatebox{90}{\underline{Stability}} \\
\midrule
NN-all & .011 & \textit{.000} & \textit{.000} & \textbf{.006} & \textit{.000} & .003 & \textit{.000} & \textbf{0} & .042 & \textit{.000} & \textit{.000} & \textit{.000} & .032 & .055 \\
NN-root & .030 & .001 & .024 & .180 & .012 & .029 & .022 & \textit{.000} & .039 & \textit{.000} & \textit{.000} & \textit{.000} & .033 & .056 \\
Min & \textbf{.005} & \textbf{0} & \textit{.000} & .024 & \textbf{0} & .001 & \textit{.000} & \textbf{0} & .009 & \textbf{0} & \textbf{0} & \textbf{0} & .006 & .031 \\
Max & .431 & .136 & .375 & .940 & .688 & .553 & .203 & .078 & .520 & .242 & .311 & .327 & .487 & .742 \\
\midrule
Borda & .016 & .002 & .006 & .113 & \textbf{0} & .003 & .013 & \cellcolor{green!25}\textbf{0} & .027 & \textit{.000} & \textit{.000} & \textit{.000} & .019 & .029 \\
EPH & .032 & \textit{.000} & .002 & .234 & .003 & \cellcolor{green!25}\textit{.000} & .001 & \textbf{0} & .060 & \textit{.000} & \textit{.000} & \textit{.000} & .049 & .067 \\
SNTV & .090 & \cellcolor{green!25}\textbf{0} & .103 & .559 & .011 & .183 & .108 & .045 & .054 & \textit{.000} & .048 & .049 & \cellcolor{green!25}\textbf{0} & .007 \\
Bloc & .031 & \textit{.000} & .001 & .223 & .003 & \cellcolor{green!25}\textbf{0} & \cellcolor{green!25}\textbf{0} & \cellcolor{green!25}\textbf{0} & .059 & \textit{.000} & \textit{.000} & \textit{.000} & .048 & .073 \\
\midrule
STV & .041 & \cellcolor{green!25}\textbf{0} & .039 & .373 & .002 & .085 & .032 & \textbf{0} & \textbf{0} & \textit{.000} & \textit{.000} & \textit{.000} & \cellcolor{green!25}\textbf{0} & \textit{.000} \\
PAV & .035 & .001 & .002 & .268 & .003 & \cellcolor{green!25}\textbf{0} & .005 & \textbf{0} & .064 & \cellcolor{green!25}\textbf{0} & \cellcolor{green!25}\textbf{0} & \textbf{0} & .052 & .064 \\
MES & .039 & .001 & .003 & .301 & .003 & .002 & .008 & \textbf{0} & .071 & \cellcolor{green!25}\textbf{0} & \cellcolor{green!25}\textbf{0} & \textbf{0} & .057 & .066 \\
CC & .179 & .036 & .154 & .719 & .041 & .291 & .156 & .060 & .264 & \cellcolor{green!25}\textbf{0} & .077 & .080 & .199 & .246 \\
seq-CC & .167 & .032 & .148 & .698 & .032 & .242 & .155 & .060 & .248 & \cellcolor{green!25}\textbf{0} & .072 & .072 & .184 & .224 \\
lex-CC & .054 & .006 & .009 & .402 & .003 & \textbf{0} & .027 & \textbf{0} & .093 & \textbf{0} & \textbf{0} & \textbf{0} & .076 & .085 \\
Monroe & .115 & .007 & .084 & .599 & .035 & .188 & .072 & \cellcolor{green!25}\textbf{0} & .174 & \cellcolor{green!25}\textbf{0} & .003 & .004 & .149 & .181 \\
Greedy M. & .054 & .002 & .021 & .400 & .004 & .010 & .026 & \cellcolor{green!25}\textbf{0} & .086 & \cellcolor{green!25}\textbf{0} & \textbf{0} & \textbf{0} & .070 & .086 \\
\midrule
MAV & .148 & .028 & .128 & .732 & .063 & .223 & .104 & \textbf{0} & .178 & .020 & .026 & .026 & .149 & .250 \\
RSD & .096 & .010 & .068 & .571 & .024 & \textbf{0} & .046 & \textbf{0} & .119 & .023 & .024 & .025 & .100 & .234 \\
Random & .226 & .068 & .189 & .833 & .084 & .354 & .180 & .071 & .285 & .048 & .119 & .123 & .222 & .359 \\
\bottomrule
\end{tabular}
\caption{Average Axiom Violation Rate for 6 alternatives and $1 \leq k < 6$ winners across all preferences.}
\end{table}

\begin{figure}[ht]
    \centering
    \includegraphics[width=\linewidth]{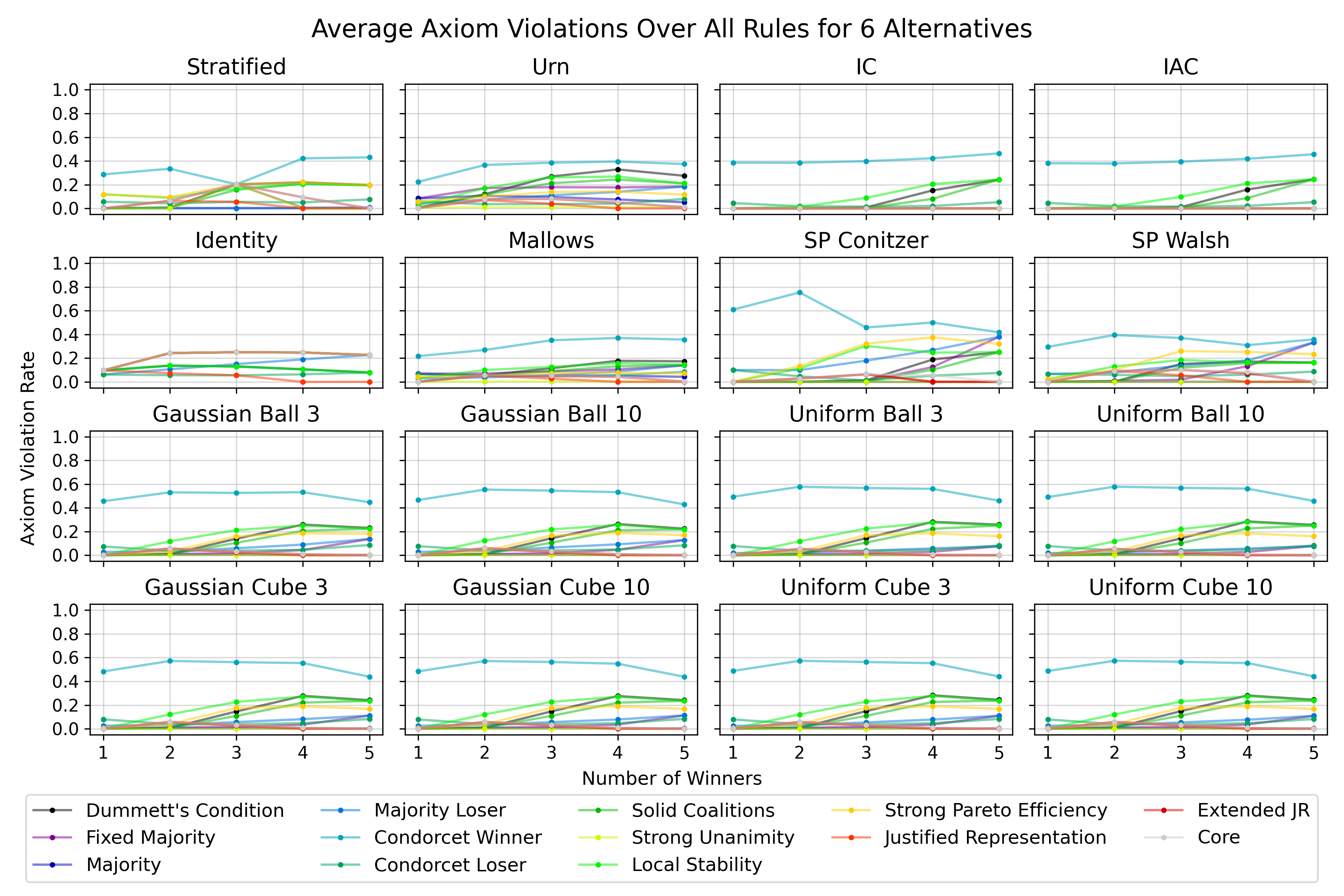}
    \caption{
    Violation rates of each axiom for every preference distribution, averaged over all rules. This shows trends common across all rules in the relationships between axioms and preference distributions.
    }
    \label{fig:violations_by_rule-m=6}
\end{figure}

\begin{table}[ht]
\centering
\fontsize{7pt}{9pt}\selectfont
\setlength{\tabcolsep}{4.6pt}
\renewcommand{\arraystretch}{1.05}
\begin{tabular}{@{}lcccccccccccccc@{}}
\toprule
 & \rotatebox{90}{Borda} & \rotatebox{90}{EPH} & \rotatebox{90}{SNTV} & \rotatebox{90}{Bloc} & \rotatebox{90}{STV} & \rotatebox{90}{PAV} & \rotatebox{90}{MES} & \rotatebox{90}{CC} & \rotatebox{90}{seq-CC} & \rotatebox{90}{lex-CC} & \rotatebox{90}{Monroe} & \rotatebox{90}{Greedy M.} & \rotatebox{90}{MAV} & \rotatebox{90}{RSD} \\
\midrule
Borda & 0 & -- & -- & -- & -- & -- & -- & -- & -- & -- & -- & -- & -- & -- \\
EPH & \cellcolor{blue!19} .238 & 0 & -- & -- & -- & -- & -- & -- & -- & -- & -- & -- & -- & -- \\
SNTV & \cellcolor{blue!34} .427 & \cellcolor{blue!30} .385 & 0 & -- & -- & -- & -- & -- & -- & -- & -- & -- & -- & -- \\
Bloc & \cellcolor{blue!19} .239 & \cellcolor{blue!1} .016 & \cellcolor{blue!30} .382 & 0 & -- & -- & -- & -- & -- & -- & -- & -- & -- & -- \\
STV & \cellcolor{blue!22} .278 & \cellcolor{blue!26} .330 & \cellcolor{blue!22} .280 & \cellcolor{blue!26} .330 & 0 & -- & -- & -- & -- & -- & -- & -- & -- & -- \\
PAV & \cellcolor{blue!19} .243 & \cellcolor{blue!3} .047 & \cellcolor{blue!31} .394 & \cellcolor{blue!4} .058 & \cellcolor{blue!26} .334 & 0 & -- & -- & -- & -- & -- & -- & -- & -- \\
MES & \cellcolor{blue!20} .255 & \cellcolor{blue!8} .107 & \cellcolor{blue!30} .376 & \cellcolor{blue!9} .117 & \cellcolor{blue!27} .342 & \cellcolor{blue!6} .076 & 0 & -- & -- & -- & -- & -- & -- & -- \\
CC & \cellcolor{blue!45} .569 & \cellcolor{blue!35} .449 & \cellcolor{blue!43} .548 & \cellcolor{blue!36} .453 & \cellcolor{blue!43} .542 & \cellcolor{blue!34} .429 & \cellcolor{blue!36} .460 & 0 & -- & -- & -- & -- & -- & -- \\
seq-CC & \cellcolor{blue!43} .538 & \cellcolor{blue!36} .453 & \cellcolor{blue!34} .437 & \cellcolor{blue!36} .460 & \cellcolor{blue!42} .533 & \cellcolor{blue!34} .435 & \cellcolor{blue!31} .399 & \cellcolor{blue!48} .608 & 0 & -- & -- & -- & -- & -- \\
lex-CC & \cellcolor{blue!24} .301 & \cellcolor{blue!10} .126 & \cellcolor{blue!33} .421 & \cellcolor{blue!10} .135 & \cellcolor{blue!29} .371 & \cellcolor{blue!6} .085 & \cellcolor{blue!9} .115 & \cellcolor{blue!31} .397 & \cellcolor{blue!33} .424 & 0 & -- & -- & -- & -- \\
Monroe & \cellcolor{blue!38} .484 & \cellcolor{blue!29} .363 & \cellcolor{blue!39} .490 & \cellcolor{blue!29} .366 & \cellcolor{blue!36} .460 & \cellcolor{blue!27} .343 & \cellcolor{blue!30} .375 & \cellcolor{blue!8} .111 & \cellcolor{blue!45} .568 & \cellcolor{blue!27} .344 & 0 & -- & -- & -- \\
Greedy M. & \cellcolor{blue!25} .318 & \cellcolor{blue!16} .210 & \cellcolor{blue!31} .394 & \cellcolor{blue!17} .217 & \cellcolor{blue!29} .374 & \cellcolor{blue!14} .187 & \cellcolor{blue!12} .155 & \cellcolor{blue!37} .470 & \cellcolor{blue!29} .372 & \cellcolor{blue!16} .202 & \cellcolor{blue!31} .397 & 0 & -- & -- \\
MAV & \cellcolor{blue!48} .602 & \cellcolor{blue!47} .589 & \cellcolor{blue!53} .673 & \cellcolor{blue!47} .588 & \cellcolor{blue!48} .604 & \cellcolor{blue!46} .585 & \cellcolor{blue!48} .600 & \cellcolor{blue!28} .352 & \cellcolor{blue!62} .781 & \cellcolor{blue!45} .567 & \cellcolor{blue!28} .350 & \cellcolor{blue!49} .619 & 0 & -- \\
RSD & \cellcolor{blue!38} .476 & \cellcolor{blue!36} .454 & \cellcolor{blue!45} .565 & \cellcolor{blue!36} .453 & \cellcolor{blue!40} .508 & \cellcolor{blue!36} .456 & \cellcolor{blue!36} .458 & \cellcolor{blue!49} .624 & \cellcolor{blue!48} .608 & \cellcolor{blue!37} .472 & \cellcolor{blue!44} .557 & \cellcolor{blue!37} .472 & \cellcolor{blue!49} .614 & 0 \\
Random & \cellcolor{blue!56} .700 & \cellcolor{blue!56} .700 & \cellcolor{blue!56} .700 & \cellcolor{blue!56} .700 & \cellcolor{blue!56} .700 & \cellcolor{blue!56} .700 & \cellcolor{blue!56} .700 & \cellcolor{blue!56} .700 & \cellcolor{blue!56} .700 & \cellcolor{blue!56} .700 & \cellcolor{blue!56} .700 & \cellcolor{blue!56} .700 & \cellcolor{blue!56} .700 & \cellcolor{blue!56} .700 \\
Min & \cellcolor{blue!11} .142 & \cellcolor{blue!18} .236 & \cellcolor{blue!35} .439 & \cellcolor{blue!18} .230 & \cellcolor{blue!22} .283 & \cellcolor{blue!20} .252 & \cellcolor{blue!21} .272 & \cellcolor{blue!43} .542 & \cellcolor{blue!44} .560 & \cellcolor{blue!24} .310 & \cellcolor{blue!36} .457 & \cellcolor{blue!26} .335 & \cellcolor{blue!45} .564 & \cellcolor{blue!38} .476 \\
Max & \cellcolor{blue!75} .941 & \cellcolor{blue!74} .933 & \cellcolor{blue!69} .870 & \cellcolor{blue!74} .937 & \cellcolor{blue!73} .914 & \cellcolor{blue!73} .924 & \cellcolor{blue!73} .916 & \cellcolor{blue!64} .802 & \cellcolor{blue!60} .762 & \cellcolor{blue!72} .900 & \cellcolor{blue!68} .855 & \cellcolor{blue!71} .896 & \cellcolor{blue!64} .810 & \cellcolor{blue!67} .846 \\
NN-all & \cellcolor{blue!10} .137 & \cellcolor{blue!17} .221 & \cellcolor{blue!35} .445 & \cellcolor{blue!17} .215 & \cellcolor{blue!23} .291 & \cellcolor{blue!19} .238 & \cellcolor{blue!20} .260 & \cellcolor{blue!43} .544 & \cellcolor{blue!44} .557 & \cellcolor{blue!23} .298 & \cellcolor{blue!36} .459 & \cellcolor{blue!26} .327 & \cellcolor{blue!45} .564 & \cellcolor{blue!37} .472 \\
NN-root & \cellcolor{blue!26} .327 & \cellcolor{blue!27} .347 & \cellcolor{blue!39} .499 & \cellcolor{blue!27} .344 & \cellcolor{blue!30} .377 & \cellcolor{blue!28} .353 & \cellcolor{blue!29} .367 & \cellcolor{blue!41} .523 & \cellcolor{blue!49} .622 & \cellcolor{blue!31} .390 & \cellcolor{blue!32} .402 & \cellcolor{blue!32} .402 & \cellcolor{blue!35} .446 & \cellcolor{blue!37} .472 \\
\bottomrule
\end{tabular}

\caption{Difference between rules for 6 alternatives with $1 \leq k < 6$ averaged over all preference distributions.}
\label{tab:rule_distance_heatmap-m=[6]-pref_dist=all}
\end{table}

\end{document}